\documentclass[runningheads]{llncs}

\usepackage{eccv}

\usepackage{eccvabbrv}

\usepackage{graphicx}
\usepackage{booktabs}
\usepackage{algorithm}
\usepackage{algpseudocode}
\usepackage{caption}
\usepackage{varwidth}
\usepackage{wrapfig}

\usepackage[accsupp]{axessibility}  
\usepackage{multirow}
\newcommand{\real}{\mathbb{R}}

\newcommand{\loss}{\mathcal{L}}

\newcommand{\w}{\mathcal{W}}

\newcommand{\generator}{\mathbf{G}}
\newcommand{\rgb}{\mathcal{C}}
\newcommand{\ray}{\mathbf{r}}
\newcommand{\origin}{\mathbf{o}}
\newcommand{\density}{\mathbf{\sigma}}
\newcommand{\col}{c}
\newcommand{\direction}{\mathbf{d}}
\newcommand{\triplane}{\text{T}}
\newcommand{\model}{\text{Export3D}}

\usepackage[pagebackref,breaklinks,colorlinks,citecolor=eccvblue]{hyperref}

\begin{document}

\title{Learning to Generate Conditional Tri-plane for 3D-aware Expression Controllable Portrait Animation} 

\titlerunning{\model: 3D-aware Expression Controllable Portrait Animation}

\author{Taekyung Ki\inst{1}$^\star$ \and
Dongchan Min\inst{2} \and
Gyeongsu Chae\inst{1}}

\authorrunning{T. Ki et al.}

\institute{DeepBrainAI Inc., South Korea \\ \and
Graduate School of AI, KAIST, South Korea \\
\email{taek@deepbrain.io} ~ \email{alsehdcks95@kaist.ac.kr} ~ \email{gc@deepbrain.io} \\
\url{https://export3d.github.io}}
\def\thefootnote{$\star$}\footnotetext{The initial part of this work was done at AITRICS.}\def\thefootnote{\arabic{footnote}}

\maketitle

\begin{abstract}
In this paper, we present Export3D, a one-shot 3D-aware portrait animation method that is able to control the facial expression and camera view of a given portrait image.
To achieve this, we introduce a tri-plane generator with an effective expression conditioning method, which directly generates a tri-plane of 3D prior by transferring the expression parameter of 3DMM into the source image. The tri-plane is then decoded into the image of different view through a differentiable volume rendering.
Existing portrait animation methods heavily rely on image warping to transfer the expression in the motion space, challenging on disentanglement of appearance and expression. In contrast, we propose a contrastive pre-training framework for appearance-free expression parameter, eliminating undesirable appearance swap when transferring a cross-identity expression. Extensive experiments show that our pre-training framework can learn the appearance-free expression  representation hidden in 3DMM, and our model can generate 3D-aware expression controllable portrait images without appearance swap in the cross-identity manner.

\keywords{Portrait Image Animation \and Facial Expression Control \and 3D-aware Synthesis}
\end{abstract}    
\section{Introduction} \label{sec:intro}
Portrait image animation aims to generate a video of a given source identity with the driving motion. It has received a lot of attention due to the potential of virtual human services, such as cross-lingual film dubbing \cite{stylelipsync, videoretalking}, virtual avatar chatting \cite{styletalker, sadtalker}, and video conferencing \cite{osfv, live3d}. In these scenarios, it is essential to transfer the facial expression (e.g., eye-blinking, lip motion, etc.) from different person, i.e., \textit{cross-identity transfer}, while preserving the source identity. However, it is challenging due to the ambiguity between appearance and expression \cite{ambiguity} and the lack of paired data (e.g., different faces with the same expression) for disentanglement representation learning \cite{dpe}.

Most 2D-based methods rely on image warping \cite{fomm, tpsmm, styleheat, lia, mcnet}, which warps the source image to the driving image by estimating the motion between them. To impose a bottleneck for the motion representation, they encode the motion into the difference between sparse key-points \cite{fomm, tpsmm, mcnet} or latent codes \cite{lia}, which are trained in an unsupervised manner. However, in this scenario, the facial expressions are encoded into the motion space as well, in terms of local motion, which tends to be neglected due to the relatively large head motions. Furthermore, since the facial expression and the appearance are highly entangled in the image space, cross-identity expression transfer often involves the source appearance change.
DPE \cite{dpe} tackles this entanglement issue by proposing a self-supervised disentanglement learning framework based on cycle-consistency learning \cite{cyclegan}. However, it shows temporal inconsistency in the generated video due to its instability of cycle-consistency learning.

Another line of works \cite{hidenerf, otavatar, nofa, goavatar} explores facial expression control in 3D space using the neural radiance fields (NeRFs) \cite{nerf}. They leverage pre-trained latent representation of 3D GAN \cite{eg3d} for 3D facial prior where they design the expression in terms of latent code \cite{otavatar, nofa} or predict deformation field \cite{nerfies} to deform the well-constructed 3D representation, such as tri-plane \cite{hidenerf, otavatar, nofa, goavatar}.
However, the latent code cannot faithfully reconstruct the source identity \cite{otavatar}, and the point-wise deformation fields to those 3D representations yield video-level artifacts, such as flickers \cite{hidenerf}.

In this paper, we address the appearance-expression entanglement issue by proposing a contrastive pre-training framework over video datasets that produces appearance-free facial expressions with an orthogonal structure. Armed with this representation, we build a one-shot 3D-aware portrait image animation method, namely $\model$, which controls the facial expression and 3D camera view of a given source image without appearance swap. To achieve this, we design a generator architecture consisting of vision transformer (ViT) and convolution layers \cite{live3d, vit, dit} that directly generates the tri-planes from the source image and driving expression parameters. Instead of predicting the deformation fields for the expression, we introduce an expression adaptive layer normalization (EAdaLN) which can effectively transfer the driving expression to the source image. The main contributions of this work are summarized as follows:
\begin{itemize}
    \item We present \textbf{$\model$}, a one-shot \textbf{3D}-aware \textbf{port}rait image animation method that can explicitly control the facial \textbf{ex}pression and camera view of the source image only using the expression and camera parameters.
    \item We propose a \textbf{contrastive pre-training framework for the appearance-free facial expression} distilled from the 3DMM parameters where they form an orthogonal structure for different facial expressions.
    \item Extensive experiments demonstrate that our pre-training framework can learn the appearance-free expression, which enables our method to transfer the cross-identity expression without undesirable appearance swap.
\end{itemize}
\section{Related Works}
\label{sec:related}
\subsection{3D-aware Image Synthesis} \label{sec:3d_aware}
3D-aware image synthesis aims to generate images with explicit camera pose control \cite{graf, eg3d, stylenerf, gram, pigan, gramhd}. This is achieved by conditioning the camera pose parameter into generative features, which are then rendered into an RGB image through differentiable volume rendering \cite{neuralvolumes, nerf, nerfies, mipnerf}. This rendering technique has integrated with adversarial learning \cite{gan, graf, eg3d, stylenerf, gram, pigan, gramhd} to learn 3D view consistency from the unposed dataset. GRAM \cite{gram} generates a multi-view consistent image by learning the radiance field on a set of 2D surface manifolds. AniFaceGAN \cite{anifacegan} further learns the deformation fields \cite{nerfies} for the facial expression on these manifolds \cite{gram} for explicit facial expression control. EG3D \cite{eg3d} introduces a tri-plane representation that provides a strong 3D position encoding with neural volume rendering and become the one of the most prominent representation in this field. However, these methods generate portrait images from noise, requiring further process for real image manipulation.

Relying on the generateive power of EG3D, several works \cite{hfgi3d, spi, next3d, otavatar, 3dpose, triplanenet, goae, live3d, nofa} extend 2D GAN-inversion \cite{im2st, im2st++, e4e, psp} methods, which is challenging due to the multi-view consistency for a single-view image. Specifically, based on facial symmetry, SPI \cite{spi} utilizes horizontally flipped images for pseudo supervision to the occluded facial region. However, it requires multi-stage latent code optimizations. GOAE \cite{goae} proposes an encoder-based inversion for EG3D which enhances multi-view consistency via an occlusion-aware tri-plane mixing module. Live3DPortrait \cite{live3d} can reconstruct multi-view consistent portrait images by leveraging the synthetic data of pre-trained EG3D to provide multi-view supervision. However, these methods cannot explicitly manipulate the expression of the source image.

We propose a tri-plane generator architecture that can generate the tri-plane of a given source image with explicit expression control. Inspired by \cite{live3d, dit}, we design this generator with ViT and convolution layers \cite{vit}, and directly inject expression parameters into the tri-plane generating process through the expression adaptive layer normalization (EAdaLN). By leveraging the strong power of NeRF \cite{nerf, eg3d, live3d, nofa, next3d}, we decode the generated tri-plane into multi-view images with explicit expression manipulation.

\subsection{Portrait Image Animation} \label{sec:portrait_animation}
Portrait image animation, or face reenactment, is a task that animates a given source image according to the input driving condition, either audio \cite{wav2lip, stylelipsync, styletalker, sadtalker, styletalk, adnerf} or image \cite{fomm, tpsmm, osfv, lia, styleheat}. Specifically, image-driven methods transfer the motion of the driving image into the source image by learning the motion between them. Most works \cite{fomm, tpsmm, osfv} use facial key-points as a pivot representation to be aware of motion via the key-point displacement. FOMM \cite{fomm} estimates facial key-points in an unsupervised manner, approximating the motion through the first-order Taylor expansion. LIA \cite{lia} encodes a motion in terms of latent codes by introducing an orthonormal basis as a motion dictionary. However, the local motion (e.g., facial expression) and the global motion (e.g., head motion) are still entangled in those representations. DPE \cite{dpe} proposes a bidirectional cyclic training strategy to decouple the pose and expression within the latent codes, while it produces video-level artifacts due to the instability of the cycle-consistency learning.

To explicitly control the facial expression, several works leverage the expression parameters of 3D morphable models (3DMM) \cite{3dmm} in 2D \cite{pechead, styleheat} or 3D spaces \cite{otavatar, goavatar, hidenerf}. StyleHEAT \cite{styleheat} uses 3DMM to warp 2D spatial features of pre-trained StyleGAN2 \cite{stylegan2} while yielding texture sticking.
OTAvatar \cite{otavatar} proposes a one-shot test-time optimization method that optimizes identity codes of a single source image and learns expression-aware motion latent codes in the latent space of pre-trained EG3D.
HiDe-NeRF \cite{hidenerf} and NOFA \cite{nofa} take a different way by predicting an expression-aware deformation field \cite{nerfies} that deforms the tri-plane \cite{eg3d} reconstructed from the source image.

Our method belongs to image-driven approaches, distinguishing itself by not depending on 2D image warping or 3D deformation fields. Toward this, we propose the generator architecture that uses a source image and driving expression parameters to produce an expression-transferred tri-plane, wherein the expression parameters directly modulate the source visual features through the expression adaptive layer normalization (EAdaLN). Furthermore, we mitigate the appearance swap issue inherent in transferring other person's expression by introducing a contrastive pre-training method to obtain appearance-free expression representations. 

\begin{figure}[!t]
  \begin{center}
  \includegraphics[width=0.99\linewidth]{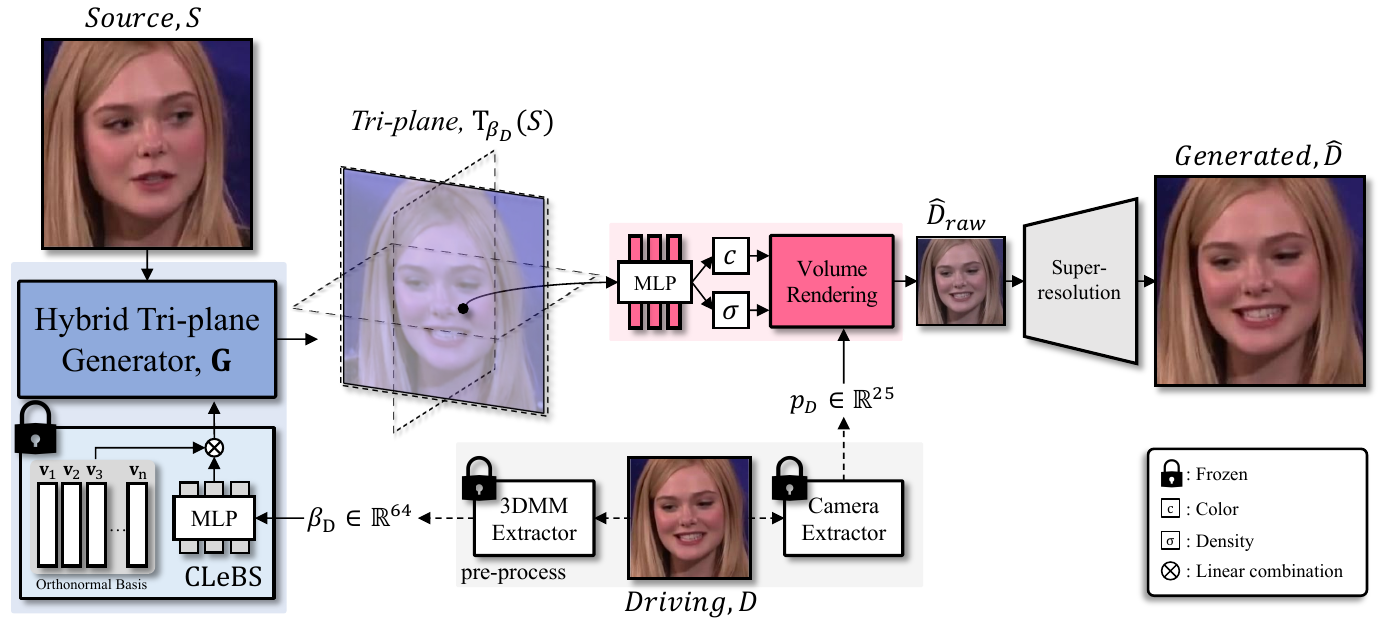}
  \caption{\small{\textbf{Training overview of \model}. We convert a source image $S\in\real^{3\times H \times W}$ into a tri-plane $T_{\beta_D}(S)$ for rich 3D priors, conditioned on an expression parameter $\beta_{D}\in \real^{64}$ from a driving image $D \in \real^{3\times H \times W}$. A differentiable volume rendering renders the tri-plane into a raw rendered image $\hat{D}_{raw}\in \real^{3\times\frac{H}{4}\times\frac{W}{4}}$ using the camera parameter $p_{D} \in \real^{25}$ of $D$, which is then super-resolved into a final image $\hat{D} \in \real^{3\times H \times W}$.}} \label{fig:overview}
  \end{center}
\end{figure}
\section{Methods} \label{sec:methods}
First of all, we formulate our portrait animation method, \textit{\model}. Given a source image  $S \in \real^{3 \times H \times W}$, our method transfers the facial expression and camera view of a driving image $D \in \real^{3\times H \times W}$ with the expression and camera parameters, respectively. We employ a tri-plane \cite{eg3d} as the intermediate feature representation, providing a strong 3D position information for differentiable volume rendering \cite{nerf, neuralvolumes}. We directly generate an expression-transferred tri-plane from the source image and the driving expression parameter \cite{3dmm, expression} through expression adaptive layer normalization (EAdaLN) (\cref{sec:hybrid}). Based on the observation that the expression parameter still contains the appearance information, we propose a pre-training framework using contrastive learning to obtain the appearance-free expression, which forms an orthogonal structure for different expressions (\cref{sec:lebs}). The expression-transferred tri-plane is rendered into a 3D-aware image through the differentiable volume rendering, and then super-resolved into the final output (\cref{sec:volume_rendering}).

\subsection{Contrastive Learned Basis Scaling (CLeBS)}\label{sec:lebs}

\begin{wrapfigure}{r}{0.48\linewidth}
    \centering
    \includegraphics[width=0.9\linewidth]{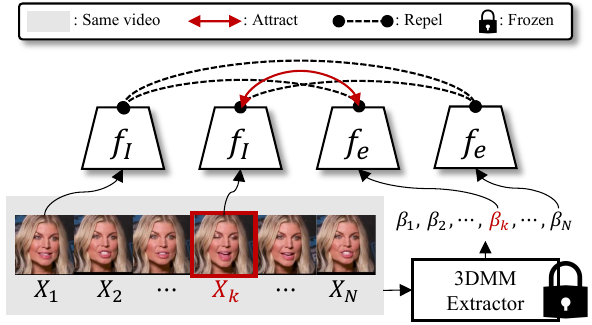}
    \caption{\small{\textbf{Contrastive pre-training framework for LeBS}. We sample the positive and the negative samples from the same video source so that those samples share the same appearance. Using contrastive learning, the encoder $f_{e}(\cdot)$ learns an appearance-free representations.}}\label{fig:pretraining_clebs}
\end{wrapfigure}

Natural speaking style comes from the the non-verbal component, such as eye blinking. To explicitly control the expression of the generated face, we utilize the expression parameter $\beta \in \real^{64}$ from the widely used 3D morphable models (3DMM) \cite{3dmm} in 3D face reconstruction. However, simply using those parameters for transferring the other person's expression fails to preserve the facial identity of the source face.

\noindent \textbf{Disentangling Expression and Appearance}. \quad  In 3DMM-based face reconstruction, the identity-appearance has been rarely explored. However, \cite{ambiguity} shows that a 3D face shape can be reconstructed only using the expression parameters not using the shape parameters, or vice versa.
We also observe that the expression parameter of 3DMM is highly entangled with the appearance (Fig. \ref{fig:tsne_exp}), resulting in an undesirable appearance swap when transferring the cross-identity expressions. We assume that the expression parameter needs to be refined to represent pure facial expressions. To address this issue, we propose a contrastive learning based pre-training framework \cite{simclr, moco, clip, styletalker} on video dataset to discard the appearance information hidden in the expression parameter. Specifically, given a video sequence $\{X_{i}\}_{i=1}^{N}$ and its corresponding expression sequence $\{\beta_{i}\}_{i=1}^{N}$, we sample an aligned image-expression pair $(X_k, \beta_k)$ for the positive and the non-aligned pairs for the negatives as illustrated in \cref{fig:pretraining_clebs}. The images and the expressions are mapped into $d$-dimensional representations, and the distance between the positive (or negative) representation pairs is minimized (or maximized) via the following contrastive objective $\loss_{cl}$:
\begin{equation}
    \loss_{cl} = -\log \left(\frac{\exp(\cos(f_{I}(X_k), f_{e}(\beta_k))/\tau)}{\sum_{j \neq k} \exp(\cos(f_{I}(X_j),f_{e}(\beta_{k}))/\tau)}\right),
    \label{eq:contra}
\end{equation}
where $f_{I}(\cdot)$ is an image encoder, $f_{e}(\cdot)$ is an expression encoder, $\tau$ is the temperature, and $\cos(\cdot, \cdot)$ is the cosine similarity, respectively. Since all samples are from the same video, they share the same appearance, thereby \cref{eq:contra} enforces the encoders to learn appearance-free expression.

Moreover, for designing the expression encoder $f_{e}(\cdot)$, we focus on the orthogonal structure of 3DMM \cite{3dmm} that controls different expressions along different orthogonal directions. To provide the appearance-free expression with the orthogonal structure, we introduce a learned orthonormal basis $V$:
\begin{equation}
    V = \{\textbf{v}_1, \textbf{v}_2, \cdots, \textbf{v}_n\} \subseteq{\real^{d}} ~~~\text{and}~~~ \langle \textbf{v}_i, \textbf{v}_j \rangle = \delta_{ij}, ~\forall i, j,
\end{equation}
spanning our new expression sub-space ($\delta_{ij}$ is the Kroneker delta function). More precisely, we convert the expression $\beta \in \real^{64}$ into the low-dimensional coefficient $\lambda = (\lambda_1, \lambda_2, \cdots, \lambda_n) \in \real^{n} ~ (n \ll 64)$ and then scales the learned orthonormal basis $V \subseteq \real^d$ to produce the appearance-free expression representation $\beta' \in \real^{d}$:
\begin{equation}
    \beta' = f_e(\beta) = \lambda_{1} \textbf{v}_1 + \lambda_{2} \textbf{v}_2 + \cdots + \lambda_{n} \textbf{v}_n \in \real^{d}.
    \label{eq:beta'}
\end{equation}
We apply QR-decomposition \cite{lia} to a learned weight ($\in \real^{d \times n}$) to explicitly compute the orthonormal basis $V \in \real^{d \times n}$. In this space, an expression is a linear combination of the basis $V = \{ \textbf{v}_i\}_{i=1}^{n}$ where the coefficient $\lambda = (\lambda_1, \lambda_2, \cdots, \lambda_n)$ is responsible for the intensity of each expression direction. We call our encoder $f_{e}(\cdot)$ a learned basis scaling (LeBS) module with contrastive pre-training (CLeBS). Once CLeBS is pre-trained with \cref{eq:contra}, no further training is required as illustrated in \cref{fig:overview} and \cref{fig:encoders}, and the image encoder $f_I (\cdot)$ is never used after then. 

\subsection{Hybrid Tri-plane Generator} \label{sec:hybrid}
We employ the tri-plane as the intermediate feature representation for 3D prior to volume rendering. Tri-plane $\triplane$ consists of features assigned on the 3 axis-aligned planes (i.e., $xy, yz, zx$ planes): 
\begin{equation}
\triplane = (\triplane_{xy}, \triplane_{yz}, \triplane_{zx}) \in \real^{3\times32\times \frac{H}{2}\times \frac{W}{2}},
\end{equation}
where $\triplane_{ij} \in \real^{32\times \frac{H}{2}\times \frac{W}{2}}$ is the 32-dimensional feature of $\frac{H}{2}\times\frac{W}{2}$ resolution on the $ij$-plane. EG3D \cite{eg3d} utilizes StyleGAN2 \cite{stylegan2} to generate the tri-plane from a noise, forming the style latent space $\w \subseteq \real^{512}$.
Several works \cite{hfgi3d, spi, next3d, otavatar, 3dpose, triplanenet, nofa} extend the 2D GAN-inversion methods \cite{im2st,im2st++, psp, e4e, pti} to 3D GAN-inversion in terms of reconstructing the tri-plane from the style latent code. \begin{wrapfigure}{r}{0.5\linewidth}
  \begin{center}
  \includegraphics[width=0.9\linewidth]{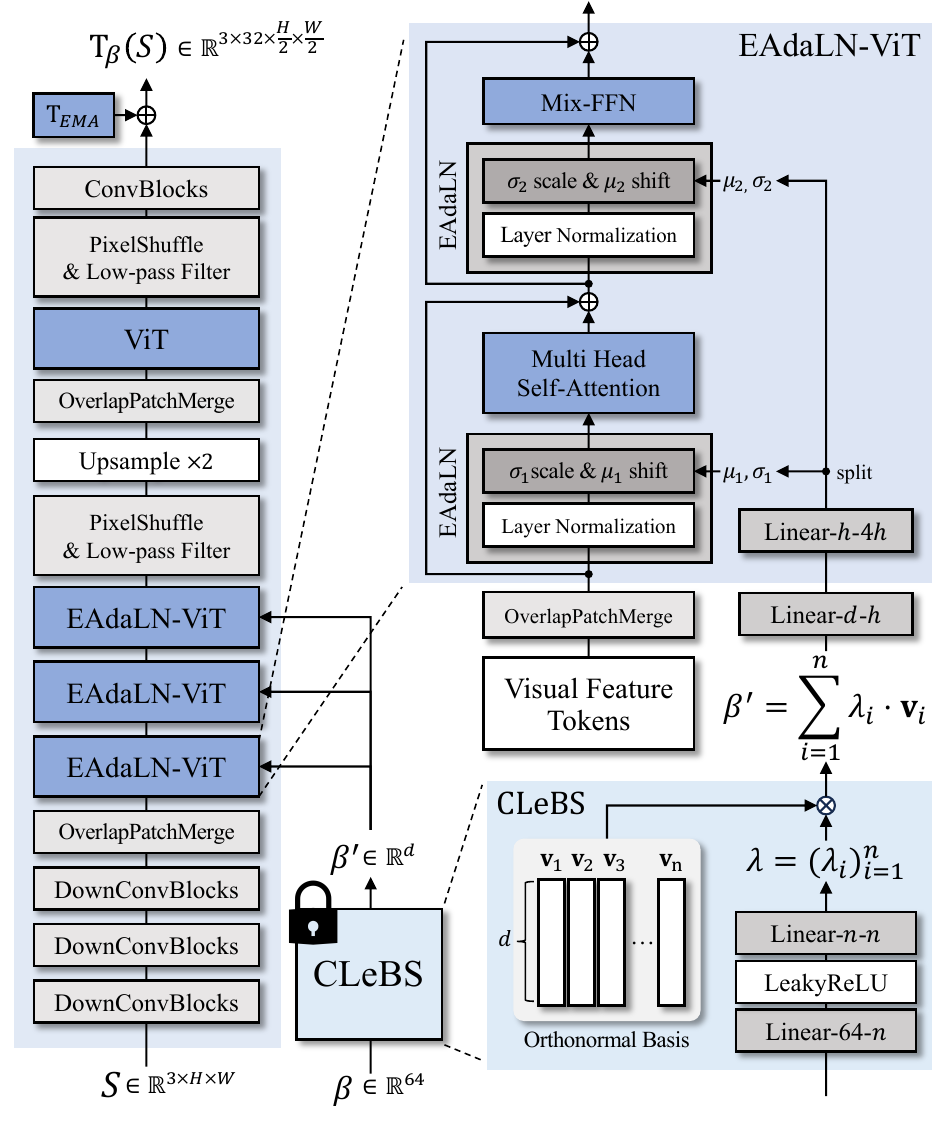}
  \caption{\small{\textbf{Hybrid tri-plane generator $\generator$ and Expression Adaptive Layer Normalization (EAdaLN)}. EAdaLN modulates the expression of $S$ using the refined expression $\beta'$ from CLeBS.}}
  \label{fig:encoders}
  \end{center}
\end{wrapfigure} However, these methods often face challenges in preserving facial identity since the style latent code lacks the capacity for encoding spatial information and person-specific visual details. We directly generate the expression-transferred tri-plane $\triplane_{\beta_{D}}(S)$ from the source $S$ and the driving expression $\beta_D \in \real^{64}$ to reconstruct the driving $D$. Inspired by Live3DPortrait \cite{live3d}, we construct the tri-plane generator with ViT and convolution \cite{vit, segformer}. Specifically, we convert $S\in \real^{3 \times H \times W}$ into a visual feature in $\real^{\frac{h}{4} \times \frac{H}{2^3} \times \frac{W}{2^3}}$ through a stack of convolutional blocks, and then merge it  into the $h$-dimensional $\frac{H \cdot W}{2^8}$ visual tokens through a overlap patch merge operator \cite{segformer}. These tokens and driving expression are processed through a conditional ViT \cite{vit, dit, segformer} blocks, namely \textit{EAdaLN-ViT}, where the expression modulates \cite{dit} the visual tokens through expression adaptive layer normalization (EAdaLN) as illustrated in \cref{fig:encoders}. EAdaLN is applied right before the multi-head self-attention (MSA) and the mix feed-forward network (Mix-FFN)\cite{segformer} of each ViT block to inject the semantic expression into the visual tokens:
\begin{equation}
    \text{EAdaLN}(x, \beta'_{D}) = \sigma(\beta'_{D}) \times \text{LN}(x) + \mu(\beta'_{D}) \in \real^{h \times \left(\frac{H\cdot W}{2^8}\right)},
\end{equation}
where $x$ is the input visual token, $\text{LN}(\cdot)$ is the layer normalization \cite{layernorm}, $\sigma(\beta'_{D})$ and $\mu(\beta'_{D})$ are the $h$-dimensional scale and shift factors computed from $\beta'_{D} = f_{e}(\beta_{D}) \in \real^d$, respectively. To efficiently propagate the visual tokens to the higher resolution, we upsample the visual tokens with pixel shuffle \cite{live3d} followed by the Gaussian low-pass filter \cite{stylegan3}. We experimentally find that the tokens and the pixel shuffle produce grid artifacts, challenging to eliminate in the image space. Employing low-pass filters effectively mitigates these artifacts by smoothing the borderline artifacts over the coordinate. Lastly, we use ViT and convolutional blocks to output the tri-plane  $\triplane_{\beta_{D}}(S)$:
\begin{equation}
    \triplane_{\beta_{D}}(S) = {\generator}(S, \beta_{D}) \in \real^{3 \times 32 \times \frac{H}{2} \times \frac{W}{2}}.
\end{equation}
Note that our method does not query the expression parameter to estimate the motion \cite{dpe, styleheat, pechead}, rather it is used as the multi-dimensional label. To stabilize the tri-plane generation, we incorporate the online exponential moving average (EMA) over tri-plane $\triplane_{EMA}$ which is added to the generated tri-plane. Please refer to supplementary materials for detailed architectures. 

\subsection{Volume Rendering and Super-resolution} \label{sec:volume_rendering}

The tri-plane can be rendered into a 2D RGB image through the differentiable volume rendering \cite{nerf, neuralvolumes, eg3d}. The expression-transferred tri-plane $\triplane_{\beta_D}(S)$ is projected onto 3 orthogonal planes ($xy, yz, zx$-planes) and then aggregated through average \cite{eg3d}:
\begin{equation}
F_{\beta_D}(S) = \frac{1}{3}(F_{\beta_D, xy}(S) + F_{\beta_D,yz}(S) + F_{\beta_D,zx}(S)), 
\end{equation}
where $F_{\beta_D, ij}(S)$ are the projected features of $T_{\beta_D}(S)$ onto the $ij$ planes. A light-weight $\text{MLP}$ assigns a color $\col$ and density $\density$ to each point $(x,y,z)$ using the aggregated feature $F_{\beta_D}(S)$: 
\begin{equation}
    \text{MLP}: F_{\beta_D}(S) \longrightarrow (\col, \density).
\end{equation}
The differentiable volume rendering \cite{nerf, eg3d, adnerf} composites each color $\col$ and density $\density$ into a RGB value $\rgb$ along the camera ray $\ray$:
\begin{equation}
    \rgb = \int_{t_n}^{t_f} \density (\ray(t)) \cdot \col(\ray(t)) \cdot T(t) dt,
    \label{eq:volume_render}
\end{equation}
where $\ray(t) = \origin + t \direction$, $t \in [t_n, t_f]$, with camera center $\origin \in \real^{3}$, viewing direction $\direction \in \real^{3}$, and $T(t)$ is the accumulation measure along the ray $\ray$ from $t_{n}$ to $t$:
\begin{equation}
    T(t) = \exp{ \left( -\int_{t_n}^{t} \density(\ray(s)) ds \right)}.
    \label{eq:transmittance}
\end{equation}
Note that the ray $\ray$ is determined by the driving camera parameter $p_{D}\in\real^{25}$ to render the generated tri-plane $\triplane_{\beta_D}(S)$ into a image of the same view with $D$. As the appearance and the expression are already encoded in the tri-plane generation, the volume rendering can determine the view-consistent images.

Directly rendering a target high-resolution image requires high computational cost. One promising approach to address this issue is to incorporate super-resolution blocks \cite{eg3d, gramhd, live3d} that upsamples the rendered image of low resolution. Following this approach, we first render a $\hat{D}_{raw} \in \real^{3 \times \frac{H}{4} \times \frac{W}{4}}$ and then apply the super-resolution to obtain the target resolution $\hat{D} \in \real^{3 \times H \times W}$, as illustrated in \cref{fig:overview}. Instead of using style-modulated convolution \cite{eg3d, live3d}, we use plane convolutional blocks for super-resolution, as we do not leverage the style latent code. Detailed architecture is provided in supplementary materials.

\section{Experiments} \label{sec:exp}
\begin{figure}[!t]
    \begin{center}
    \captionof{table}{\textbf{Quantitative comparison on VFHQ.} The best score for each metric is in \textbf{bold}. Note that we only measure CSIM \cite{arcface}, AED and APD \cite{bfm, pirenderer} for the cross-identity experiment as no ground-truth is available. \\ {${^\dagger}$: Evaluated only on the foreground facial region.}} \label{tab:quantitative}
    \resizebox{0.95\textwidth}{!}{
        \begin{tabular}{l | c c c c c c | c c c}
        \toprule
        \multicolumn{1}{c|}{\multirow{2}{*}{Model}} & \multicolumn{6}{c|}{Same-identity} & \multicolumn{3}{c}{Cross-identity} \\
        \cline{2-10}
        \multicolumn{1}{c|}{} & \textbf{PSNR} $\uparrow$ & \textbf{SSIM} $\uparrow$ & \textbf{AKD} $\downarrow$ & \textbf{CSIM} $\uparrow$ & \textbf{AED} $\downarrow$ & \textbf{APD} $\downarrow$ & \textbf{CSIM} $\uparrow$ & \textbf{AED} $\downarrow$ & \textbf{APD} $\downarrow$ \\
        \hline
        StyleHEAT \cite{styleheat}  & 14.233 & 0.428 & 30.406 & 0.464 & 0.161 & 0.139 & 0.505 & 0.242 & 0.136 \\
        DPE \cite{dpe} & 23.241 & \textbf{0.750} & 3.661 & 0.831 & 0.083 & 0.032 & 0.586 & 0.253 & 0.085 \\ 
        ROME${^\dagger}$ \cite{rome}  & 14.185 & 0.642 & 7.281 & 0.737 & 0.111 & 0.051 & 0.641 & 0.224 & 0.074 \\
        OTAvatar${^\dagger}$ \cite{otavatar} & 17.441 & 0.651 & 11.502 & 0.662 & 0.176 & 0.067 & 0.610 & 0.290 & 0.198 \\
        HiDe-NeRF${^\dagger}$ \cite{hidenerf} & 21.228 & 0.728 &  8.245 & \textbf{0.867} & 0.106 & 0.049 & \textbf{0.707} & 0.255 & \textbf{0.065} \\ 
        \hline
        \textbf{Ours} & \textbf{23.555} & 0.704 & \textbf{3.453} & 0.811 & \textbf{0.082} & \textbf{0.030} & 0.694 & \textbf{0.208} & 0.080 \\
        \bottomrule
        \end{tabular}
        }
    \end{center}
    \begin{center}
    \includegraphics[width=0.95\textwidth]{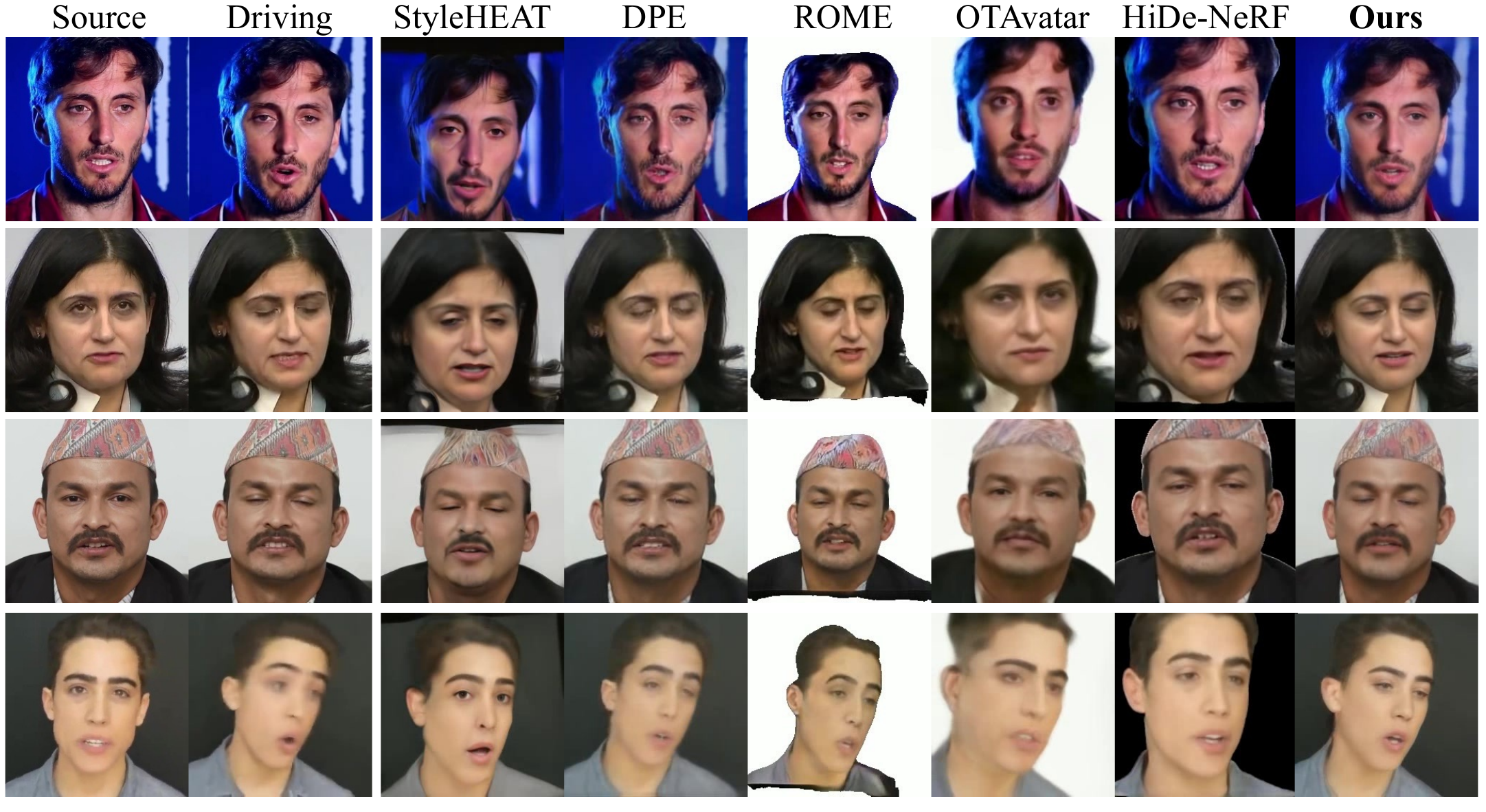}
    \captionof{figure}{\textbf{Comparison on same-identity experiments.} For a fair comparison, we follow the pre-processing strategy of each method.}\label{fig:same_identity}
    \end{center}
\end{figure}

\subsection{Dataset and Pre-processing}\label{sec:dataset}

We train our model on real video dataset VFHQ \cite{vfhq}. Following the video pre-processing strategies in \cite{fomm, stylelipsync}, we convert the original video into 25 fps and crop the facial regions of resolution $256 \times 256$, ensuring that the nose is located at the center of the image. We use a 3DMM extractor \cite{bfm} to obtain the expression parameter $\beta \in \real^{64}$. We adopt the pre-preprocesing strategy of EG3D \cite{eg3d} for the camera parameter $p \in \real^{25}$ (the concatenation of the camera intrinsic parameters in $\real^{9}$ and the inverse extrinsic parameter in $\real^{16}$). After the video pre-processing, 6196 video clips are used for training, and 50 videos are used for test. We also evaluate our model on the test dataset of TalkingHead-1KH \cite{osfv}. After the same pre-processing, remaining 20 videos of different identities are used. 

\subsection{Evaluation} \label{sec:eval}
\begin{figure}[!t]
    \begin{center}
    \captionof{table}{\textbf{Quantitative comparison on TalkingHead-1KH.} The best score for each metric is in \textbf{bold}. Note that we only measure CSIM \cite{arcface}, AED and APD \cite{bfm, pirenderer} for the cross-identity experiment as no ground-truth is available. \\ {${^\dagger}$: Evaluated only on the foreground facial region.}} \label{tab:quantitative_th1kh}
    \resizebox{0.95\textwidth}{!}{
        \begin{tabular}{l | c c c c c c | c c c}
        \toprule
        \multicolumn{1}{c|}{\multirow{2}{*}{Method}} & \multicolumn{6}{c|}{Same-identity} & \multicolumn{3}{c}{Cross-identity} \\
        \cline{2-10}
        \multicolumn{1}{c|}{} & \textbf{PSNR} $\uparrow$ & \textbf{SSIM} $\uparrow$ & \textbf{AKD} $\downarrow$ & \textbf{CSIM} $\uparrow$ & \textbf{AED} $\downarrow$ & \textbf{APD} $\downarrow$ & \textbf{CSIM} $\uparrow$ & \textbf{AED} $\downarrow$ & \textbf{APD} $\downarrow$ \\
        \hline
        StyleHEAT \cite{styleheat}  & 15.613 & 0.517 & 21.198 & 0.575 & 0.148 & 0.095 & 0.571 & 0.218 & 0.102 \\
        DPE \cite{dpe} & 23.201 & 0.786 & 4.281 & 0.807 & 0.093 & \textbf{0.029} & 0.714 & 0.216 & 0.081 \\ 
        ROME${^\dagger}$ \cite{rome}  & 15.921 & 0.695 &  13.444 & 0.726  & 0.123  & 0.062 & 0.667 & \textbf{0.201} & 0.084 \\
        OTAvatar${^\dagger}$ \cite{otavatar} & 16.952 &  0.660 &  11.615 & 0.668 & 0.181 & 0.063 & 0.682 & 0.247 & 0.150 \\
        HiDe-NeRF${^\dagger}$ \cite{hidenerf} & 19.759 & 0.729 &  5.746 & \textbf{0.843} & 0.112 & 0.043 & 0.757 & 0.232 & 0.085 \\ 
        \hline
        \textbf{Ours} & \textbf{23.239} & \textbf{0.797} & \textbf{3.581} & 0.764 & \textbf{0.092} & 0.033 & \textbf{0.772} & 0.204 & \textbf{0.076} \\
        \bottomrule
        \end{tabular}
        }
    \end{center}
    \begin{center}
        \includegraphics[width=0.95\textwidth]{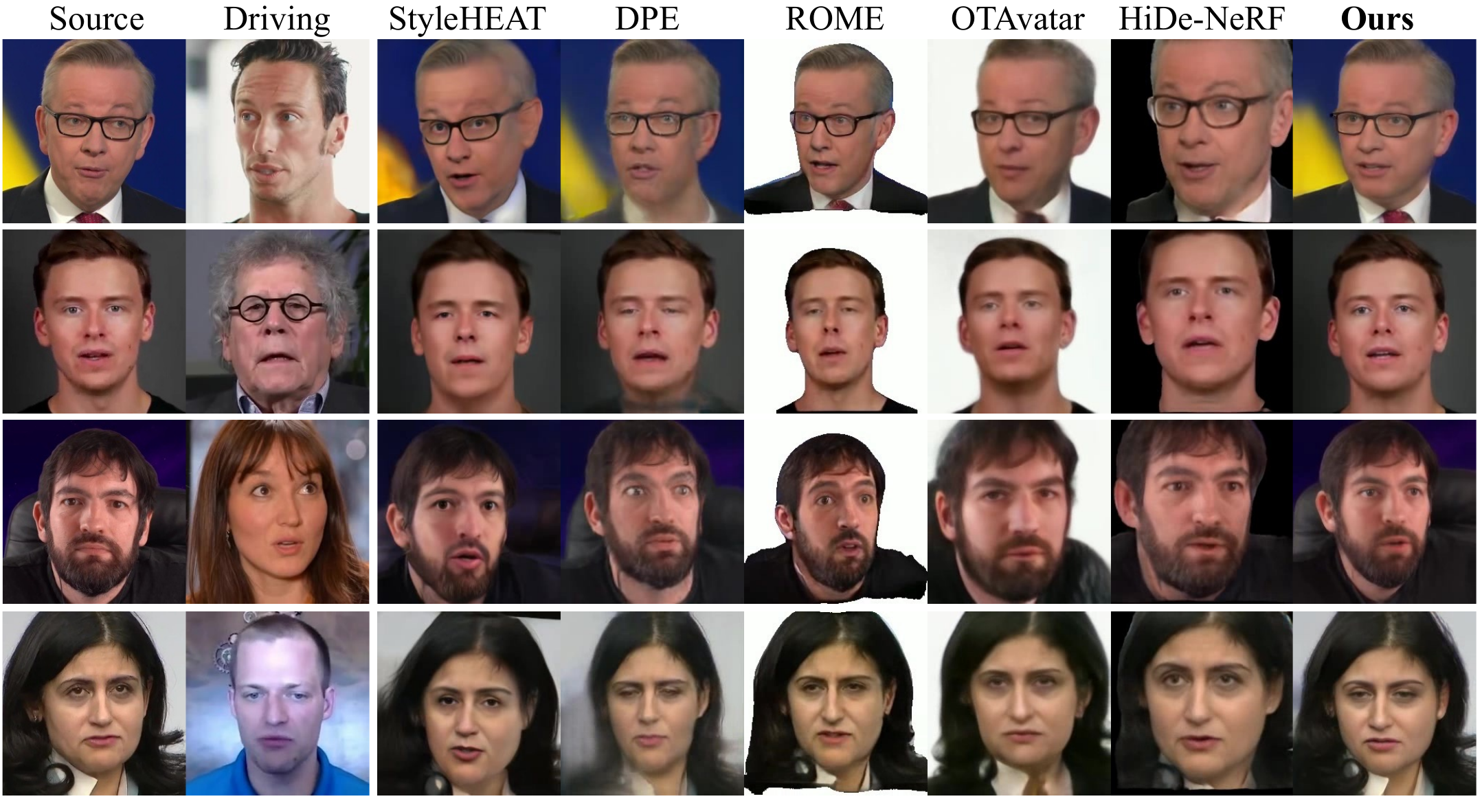}
    \caption{\textbf{Comparison on cross-identity experiments.} For a fair comparison, we follow the pre-processing strategy of each method. Notably, most portrait animation methods fail to preserve the source identity or transfer driving appearance features, such as \textit{eye shape and facial contour}, in cross-identity scenarios.}\label{fig:cross_identity}
    \end{center}
\end{figure}

We compare our model against 2D-based \cite{styleheat, dpe} and 3D-based \cite{rome, otavatar, hidenerf} image-driven portrait animation methods whose official implementations are available. \textbf{StyleHEAT} \cite{styleheat} warps the 2D spatial features of pre-trained StyleGAN2 using 3DMM parameters, \textbf{DPE} \cite{dpe} disentangles the pose and the expression in the motion latent space without using 3DMM parameters. \textbf{ROME} \cite{rome} is a mesh-based method transferring the expression and pose using 3DMM. \textbf{OTAvatar} \cite{otavatar} leverages pre-trained EG3D \cite{eg3d} by modeling head motion in terms of latent codes. \textbf{HiDe-NeRF} \cite{hidenerf} deforms the source tri-plane by predicting expression-aware deformation fields. For evaluation, we employ peak signal-to-noise ratio (\textbf{PSNR}) and structural similarity index measure (\textbf{SSIM}) for image quality, average key-point distance (\textbf{AKD}) \cite{pirenderer} for facial structure based on the 68 facial key-points, cosine similarity of identity embedding (\textbf{CSIM}) \cite{arcface} for identity preservation, average expression distance (\textbf{AED}), and average pose distance (\textbf{APD}) \cite{bfm, pirenderer} for expression transferring and pose matching. For the cross-identity experiments, we only measure CSIM, AED and APD as no ground-truth image is available.

\begin{figure}[!tb]
\begin{center}
    \captionof{table}{\textbf{Ablation studies on the expression encoding}. Same evaluation setting with \cref{tab:quantitative}. The best score for each metric is in \textbf{bold}.}\label{tab:ablation}
    \resizebox{0.95\textwidth}{!}{
    \begin{tabular}{l|cccccc|ccc}
    \toprule
    \multicolumn{1}{c|}{\multirow{2}{*}{Method}} & \multicolumn{6}{c|}{Same-identity} & \multicolumn{3}{c}{Cross-identity} \\
        \cline{2-10}
    \multicolumn{1}{c}{} & \multicolumn{1}{|c}{\textbf{PSNR}$\uparrow$} & \multicolumn{1}{c}{\textbf{SSIM}$\uparrow$} & \multicolumn{1}{c}{\textbf{AKD}$\downarrow$} & \multicolumn{1}{c}{\textbf{CSIM}$\uparrow$} & \multicolumn{1}{c}{\textbf{AED}$\downarrow$} & \multicolumn{1}{c}{\textbf{APD}$\downarrow$} & \multicolumn{1}{|c}{\textbf{CSIM}$\uparrow$} & \multicolumn{1}{c}{\textbf{AED}$\downarrow$} & \multicolumn{1}{c}{\textbf{APD}$\downarrow$} \\
        \hline
        Direct 3DMM & 23.077 & 0.688 & 3.874 & 0.789 & 0.105 & 0.044 & 0.648 & 0.209 & 0.073 \\
        E2E LeBS ($n=25$) & 23.105 & 0.672 & 3.775 & 0.745 & 0.109 & 0.040 & 0.670 & 0.218 & \textbf{0.071} \\
        E2E LeBS ($n=10$) & 23.235 & 0.676 & 3.755 & 0.751 & 0.110 & 0.038 & 0.672 & 0.238 & 0.079 \\
        E2E LeBS ($n=5$) & 22.631 & 0.646 & 4.114 & 0.658 & 0.140 & 0.046 & 0.632 & 0.246 & 0.076 \\
        \hline
        \textbf{Full (CLeBS)} & \textbf{23.555} & \textbf{0.704} & \textbf{3.453} & \textbf{0.811} & \textbf{0.082} & \textbf{0.030} & \textbf{0.694} & \textbf{0.208} & 0.080 \\
        \bottomrule
    \end{tabular}}
    \end{center}
    \begin{center}
        \includegraphics[width=0.95\textwidth]{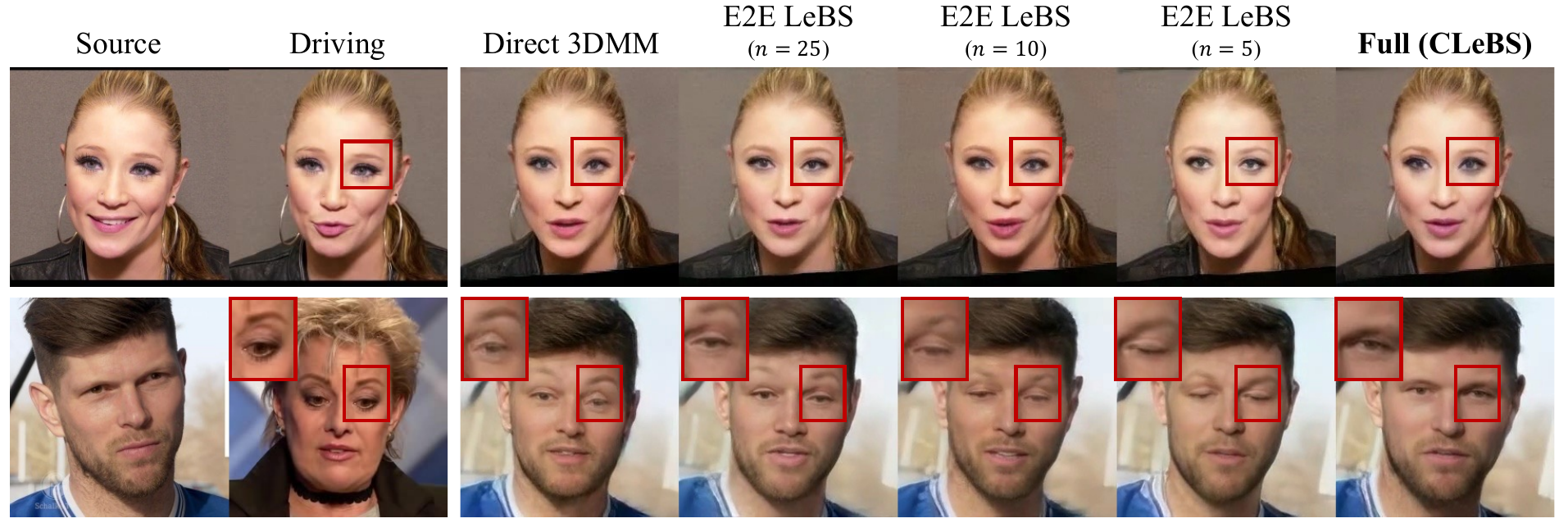}
    \captionof{figure}{\textbf{Ablation studies on the expression encoding.} Without our contrastive pre-training, the expression encoders transfer the expression together with the appearance, such as \textit{eyelids and the head size}.}\label{fig:ablation}
    \end{center}
\end{figure}

\noindent{\textbf{Same-identity experiments.}} \quad We report the same-identity transfer experiment results in \cref{tab:quantitative} and \cref{tab:quantitative_th1kh}, and illustrate the qualitative results in \cref{fig:same_identity}. For a fair comparison, ROME \cite{rome}, OTAvatar \cite{otavatar}, and HiDe-NeRF \cite{otavatar} are evaluated on the foreground facial region with different field of view. DPE \cite{dpe} shows the stable performance in the same-identity experiments with the fine-grained expression controls. Among the 3D-based methods, HiDe-NeRF \cite{hidenerf} scores the highest in the identity preservation (CSIM). Our method scores the best result in the majority of evaluation metrics. Especially, it has an advantage in expression controls (AKD and AED).

\noindent{\textbf{Cross-identity experiments.}} \quad In \cref{tab:quantitative} and \cref{tab:quantitative_th1kh}, we also conduct the cross-identity transfer experiments that transfers the expression and pose of different identity into the source identity. As illustrated in \cref{fig:cross_identity}, DPE \cite{dpe} shows visual artifacts and appearance swap, such as face contours and eye shape, due to the insufficient disentanglement of expression and pose in the motion space. HiDe-NeRF \cite{hidenerf} scores the highest identity preservation (CSIM) while un-predictable light changes are involved due to the point-wise deformation field on the tri-plane. Our method can transfer the driving expression without appearance swap and generates a video without video-level artifacts such as light changes and flickers. Please refer to our supplementary videos.

\subsection{Ablation Studies and Further Results}\label{sec:ablation}
\begin{figure}[t]
  \begin{minipage}[t]{0.485\linewidth}
    \includegraphics[width=0.95\linewidth]{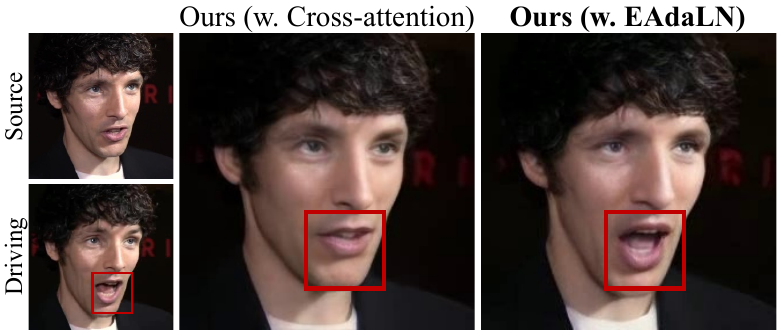}
    \caption{\textbf{Cross-attention vs. EAdaLN.}}
    \label{fig:ablation_eadaln}        
  \end{minipage}
    \begin{minipage}[b]{0.485\linewidth}
        \captionof{table}{\textbf{Ablation studies on EAdaLN.} The best score for each metric is in \textbf{bold}. We replace EAdaLN in $\generator$ with cross-attention to verify the effectiveness of EAdaLN.}\label{tab:ablation_eadaln}
    \resizebox{0.98\linewidth}{!}{
    \begin{tabular}{l|ccc|ccc}
    \toprule
    \multicolumn{1}{c|}{\multirow{2}{*}{Method}} & \multicolumn{3}{c|}{Same-identity} & \multicolumn{3}{c}{Cross-identity} \\
        \cline{2-7}
    \multicolumn{1}{c|}{} & \multicolumn{1}{c}{\textbf{CSIM}$\uparrow$} & \multicolumn{1}{c}{\textbf{AED}$\downarrow$} & \multicolumn{1}{c}{\textbf{APD}$\downarrow$} & \multicolumn{1}{|c}{\textbf{CSIM}$\uparrow$} & \multicolumn{1}{c}{\textbf{AED}$\downarrow$} & \multicolumn{1}{c}{\textbf{APD}$\downarrow$} \\
        \hline
        Ours (w. Cross-attention) & 0.678 & 0.125 & 0.042 & 0.631 & 0.271 & 0.122 \\
        \textbf{Ours (w. EAdaLN)} & \textbf{0.811} & \textbf{0.082} & \textbf{0.030} & \textbf{0.694} & \textbf{0.208} & \textbf{0.080} \\
        \bottomrule
    \end{tabular}}   
    \end{minipage}
\end{figure}

\noindent \textbf{Ablation studies on the expression encoding.} \quad In \cref{tab:ablation}, we conduct ablation studies on different expression encoding strategies. In \textbf{Direct 3DMM}, we  replace our CLeBS with fully-connected layers to directly inject the expression parameters of 3DMM through EAdaLN. As illustrated in \cref{fig:ablation}, the direct injection does not change appearance when transferring same-identity expression however, it changes appearance (e.g., eyebrows and facial contour) when transferring cross-identity expression. Furthermore, since the raw expression parameters inherently contain noise, the generated image also exhibits visual artifacts. In \textbf{E2E LeBS}, we decrease the the number of basis vectors $n$ in LeBS for appearance bottleneck to validate the proposed contrastive pre-training. Each LeBS with $n = 25, 10, 5$ is jointly trained (i.e., E2E) with entire model without any pre-training. Due to the entanglement of appearance and expression, both appearance and expression are changed as a whole as the the number of basis vector $n$ decreases. LeBS alone is insufficient for extracting appearance-free expression from the expression parameters.

\noindent \textbf{Ablation studies on EAdaLN.} \quad In \cref{tab:ablation_eadaln}, we conduct ablation studies on EAdaLN (\textbf{w. EAdaLN}) by comparing it with cross-attention (\textbf{w. Cross-attention}), which is a widely used conditioning method in transformer-based architectures \cite{transformer, dit}. Specifically, we replace all the EAdaLN blocks in $\generator$ (\cref{fig:encoders}) with cross-attention blocks. In both scenarios, CLeBS serves the refined expression $\beta'$. As shown in \cref{tab:ablation_eadaln} and \cref{fig:ablation_eadaln}, the cross-attention fails to handle the expression accurately, which verifies the effectiveness of our EAdaLN for the expression conditioning.

\begin{figure}[!tb]
    \begin{center}
        \begin{subfigure}[t]{0.42\textwidth}
            \includegraphics[width=1\textwidth]{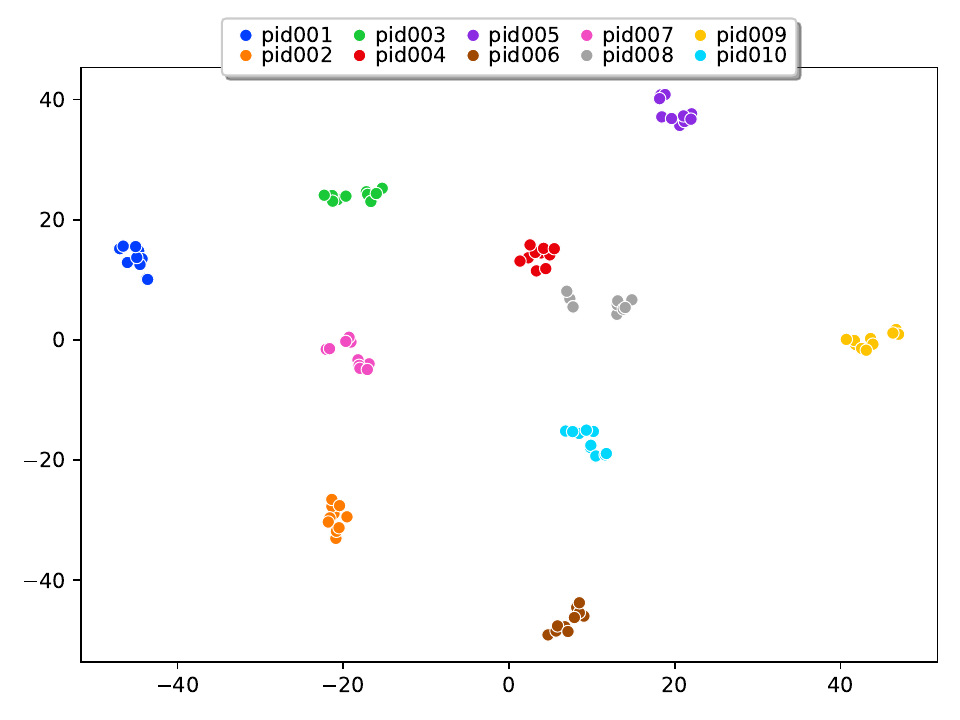} 
            \caption{\scriptsize{3DMM expression parameter $\beta$}}
            \label{fig:tsne_exp}
        \end{subfigure}
        \begin{subfigure}[t]{0.42\textwidth}
            \includegraphics[width=1\textwidth]{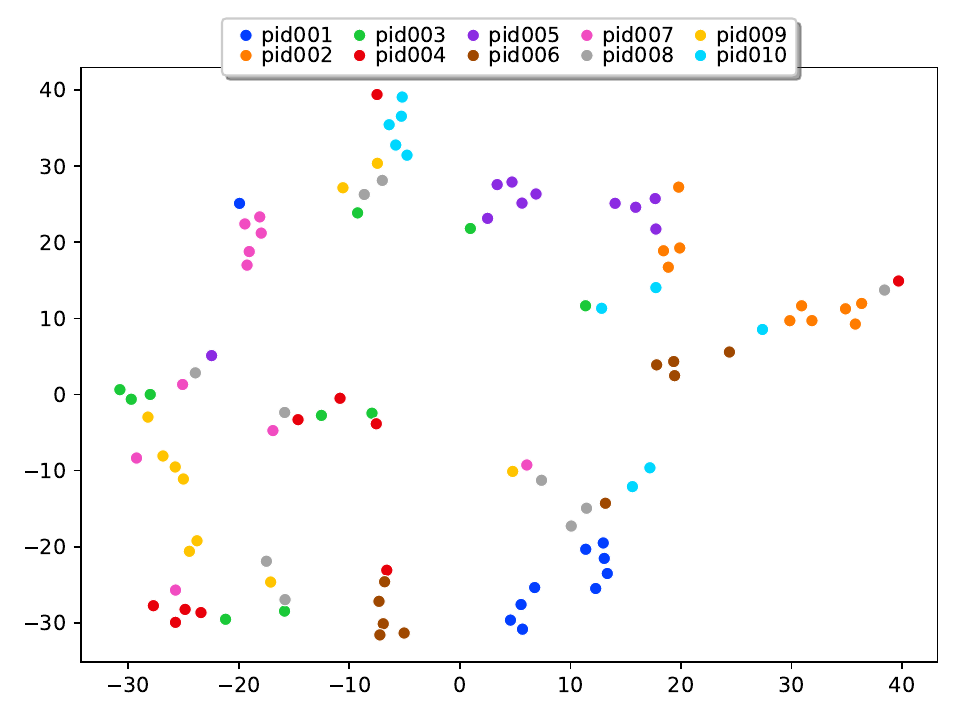}
            \caption{{\scriptsize $\beta'$ from the CLeBS.}}
            \label{fig:tsne_contastive}
        \end{subfigure}
    \caption{\textbf{Visualization of the expression parameters.} We plot t-SNE \cite{tsne} of raw 3DMM expression and our appearance-free expression parameter.}
    \label{fig:tsne}
    \end{center}
    \begin{center}
        \begin{subfigure}[t]{0.485\textwidth}
            \includegraphics[width=1\textwidth]{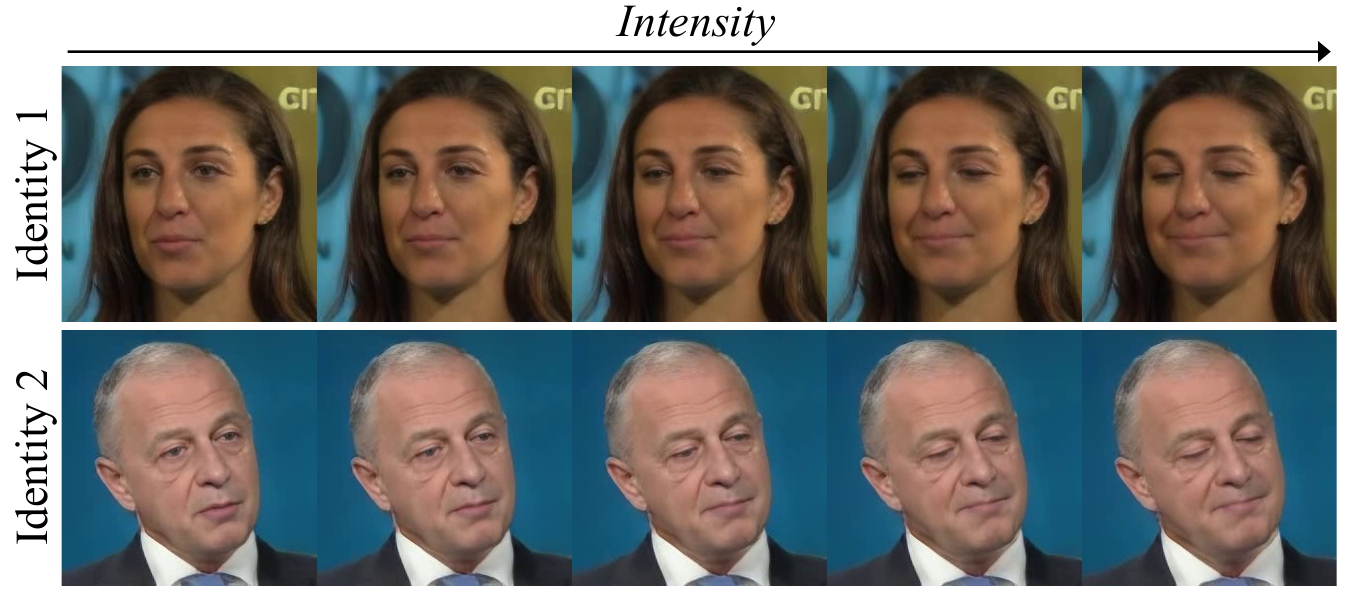} 
            \caption{Basis vector $\mathbf{v}_{4}$}
            \label{fig:linear_a}
        \end{subfigure}
        \begin{subfigure}[t]{0.485\textwidth}
            \includegraphics[width=1\textwidth]{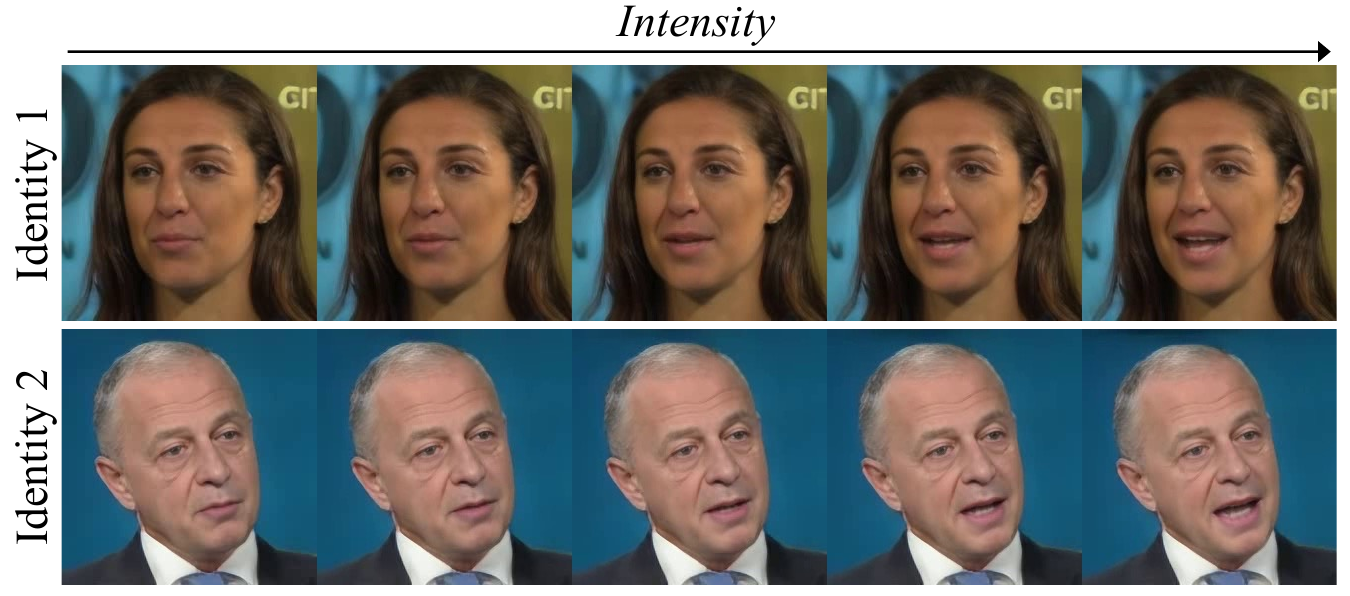}
            \caption{Basis vector $\mathbf{v}_8$}
            \label{fig:linear_b}
        \end{subfigure}
    \caption{\textbf{Linear scaling along the different basis vectors of CLeBS.} We visualize the different expression directions along the basis vectors $\mathbf{v}_{4}, \mathbf{v}_{8} \in \textbf{V}$.} \label{fig:linear}
    \end{center}
    \begin{center}
        \includegraphics[width=0.98\textwidth]{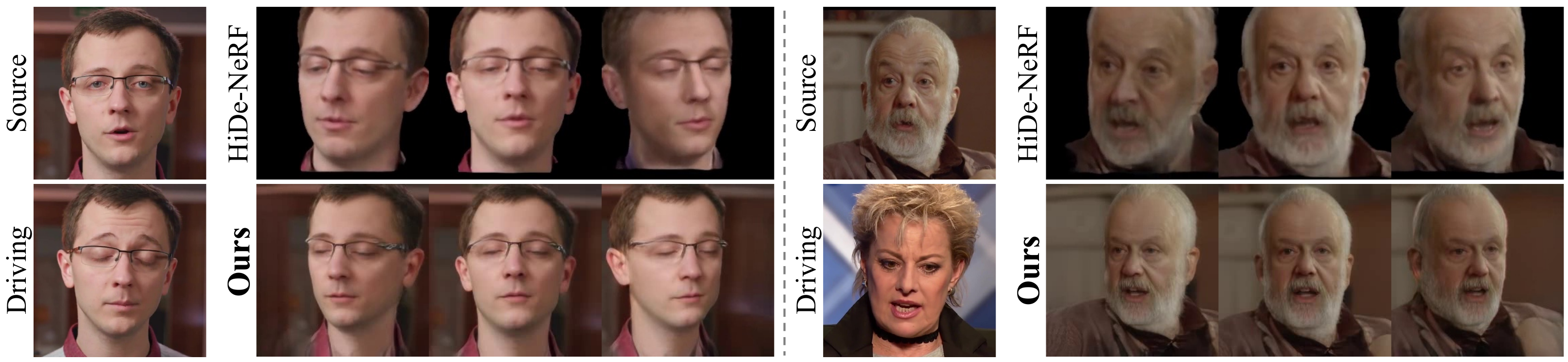}
    \caption{\textbf{Novel-view synthesis results with expression transfer.} Our method can generate more multi-view consistent images compaired to HiDe-NeRF \cite{hidenerf}.}\label{fig:vs_hidenerf}
    \end{center}
\end{figure}

\noindent \textbf{Visualization of facial expression parameters.} \quad In \cref{fig:tsne}, we sample 10 random frames from 10 different videos of distinct individual in VFHQ \cite{vfhq} and visualize the low-dimensional t-SNE \cite{tsne} results of the two expression parameters: $\beta \in \real^{64}$ and $\beta' \in \real^{d}$. In \cref{fig:tsne_exp}, the 3DMM expression parameters show strong entanglement with respective to their identities, indicating the hidden appearance information in them. On the other hand, as shown in \cref{fig:tsne_contastive}, our contrastive pre-training mitigates the entanglement, thereby resolving the appearance swap in the cross-identity expression transfer in \cref{fig:ablation}.

\noindent \textbf{Linear scaling along the orthogonal directions.} \quad In \cref{fig:linear}, we verify that $\beta' \in \real^{d}$ has the orthogonal structure where the learned basis $\mathbf{V}$ determines the different expressions even if trained in unsupervised manner and the coefficients $\{\lambda_i\}_{i=1}^{n}$ scale their intensities. Specifically, we visualize two orthogonal directions $\mathbf{v}_{4}$ and $\mathbf{v}_{8}$ and linearly scale their coefficients $\lambda_{4}$ and $\lambda_{8}$ from $1$ to $10$. As shown in \cref{fig:linear_a}, $\mathbf{v}_4$ controls eye closing and mouth closing, while \cref{fig:linear_b} illustrates that $\mathbf{v}_8$ controls mouth opening. Notably, the orthogonal basis does not influence head movements.

\noindent \textbf{Novel-view synthesis with expression transfer.} \quad In \cref{fig:vs_hidenerf}, we compare the results of novel-view synthesis with expression transfer to those of HiDe-NeRF \cite{hidenerf}. Both methods utilize the tri-plane and differentiable volume rendering to generate novel-view images. However, while HiDe-NeRF transfers the driving expression by deforming the generated tri-plane into a canonical tri-plane based on driving conditions \cite{nerfies}, our method relies on the hybrid generator $\generator$ with EAdaLN. In both same-identity and cross-identity transfer scenarios, our method synthesizes more view-consistent results, highlighting the effectiveness of our method in expression transfer without relying on deformation. Please refer to supplementary videos.

\section{Conclusion}\label{sec:conclusion}
We presented $\model$, a 3D-aware portrait image animation model that controls the facial expression and the camera view of a source image by leveraging the driving 3DMM expression and camera parameters. Since the expression parameters are still entangled with appearance information, we proposed a contrastive pre-training framework to extract appearance-free expressions from the parameters. These refined expressions are injected into our generator through expression adaptive layer normalization (EAdaLN) that produces a tri-plane of source identity and driving expression. Finally, differentiable volume rendering renders the tri-plane into 2D images of different views. Extensive experiments show that our contrastive pre-training framework removes the appearance information from the 3DMM expression parameters, enabling our model to transfer the cross-identity expressions without undesirable appearance swap.

\noindent \textbf{Limitations and future work.} \quad While our method can generate realistic face videos with driving expressions and views, it still has several limitations. First, our method cannot separately generate the foreground and background regions as the tri-plane representation construct them as a whole. Several works address this limitation by extending the tri-plane representation \cite{panohead}, restricting rendering points in the ray marching process \cite{ballgan}, or leveraging the off-the-shelf segmentation model \cite{modnet} to manually separate them \cite{hidenerf, goavatar, otavatar}. Second, our method cannot control non-facial body parts (e.g., neck and shoulders) and eye gazing as they are beyond the capability of the 3DMM parameters. We plan to address these limitations for future work. 

\noindent \textbf{Ethical consideration.} \quad Since our method can generate a realistic video using a single portrait image, it has the potential for misuse, such as fake news creations.
We have planned to attach visible and invisible watermarks to the generated videos and restrict the source identities for inference in research demonstration.

%
%
\bibliographystyle{splncs04}
\bibliography{main}
\newpage
\section{Supplementary Material}
\subsection{3D Morphable Models (3DMM).}\label{suppl_3dmm} \begin{wrapfigure}{r}{0.45\linewidth}
    \centering
    \includegraphics[width=0.99\linewidth]{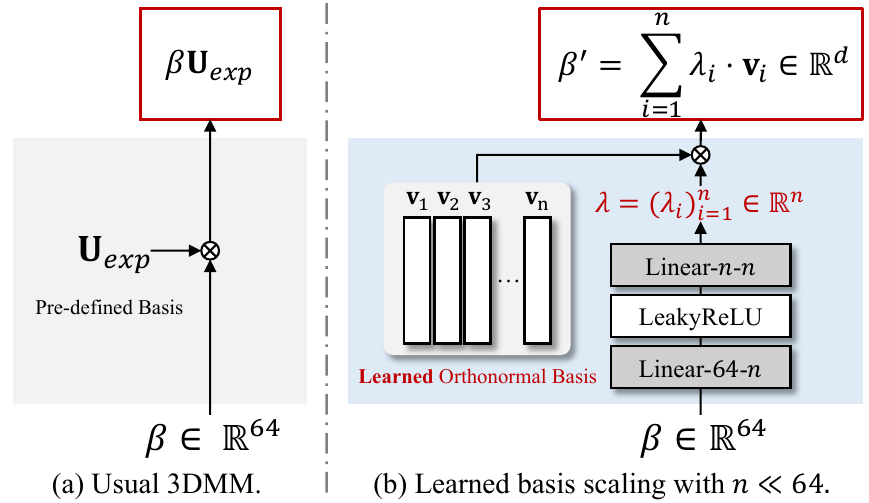}
    \caption{\small{\textbf{3DMM \cite{3dmm} vs. Leaned Basis Scaling (LeBS)}. 3DMM based method reconstructs 3D facial geometry by scaling the the pre-defined basis $\mathbf{U}_{exp}$ with expression parameters $\beta \in \real^{64}$. LeBS, on the other hand, uses the learned basis $V = \{v_{i}\}_{i=1}^{n} \subseteq \real^{d}$ which is scaled by the low-dimensional coefficients $\lambda = (\lambda_{i})_{i=1}^{n} \in \real^{n}$ ($n\ll64$).}}
    \label{fig:3dmm_vs_lebs}
\end{wrapfigure} \quad 3D Morphable Models (3DMM) \cite{3dmm} are statistical models of 3D shape and their corresponding texture. In this paper, we only consider the shape representation of 3DMM. To be specific, a face shape $\mathbf{S}$ is initialized with the average shape $\bar{\mathbf{S}}$ and further shaped by a linear combination of expression and identity as follows:
\begin{equation}
    \mathbf{S} = \bar{\mathbf{S}} + \alpha \mathbf{U}_{id} + \beta \mathbf{U}_{exp}, \label{eq:3dmm}
\end{equation}
where $\mathbf{U}_{id} \in \real^{80 \times d_{3dmm}}$, $\mathbf{U}_{exp} \in \real^{68\times d_{3dmm}}$ are the pre-defined bases of identity and expression subspaces of 3D face space, respectively. $d_{3dmm}$ is the dimension of the 3D face space. The coefficients $\alpha \in \real^{80}$ and $\beta \in \real^{64}$ determine the facial identity and expression for the face geometry reconstruction by scaling each basis vector \cite{expression}.

In this paper, we term \textbf{appearance} as the set of geometric features that determine the facial identity of a given face, such as head size, face contour, face proportion, eyebrows, eye shape, mouth shape, jaw shape, etc., and \textbf{expression} as the motion of these appearance features, such as mouth opening (closing), eye blinking, etc.

\subsection{Detailed Model Architectures.} \label{sec:suppl_model_architecture}
\begin{figure}[!t]
    \begin{center}
        \includegraphics[width=\textwidth]{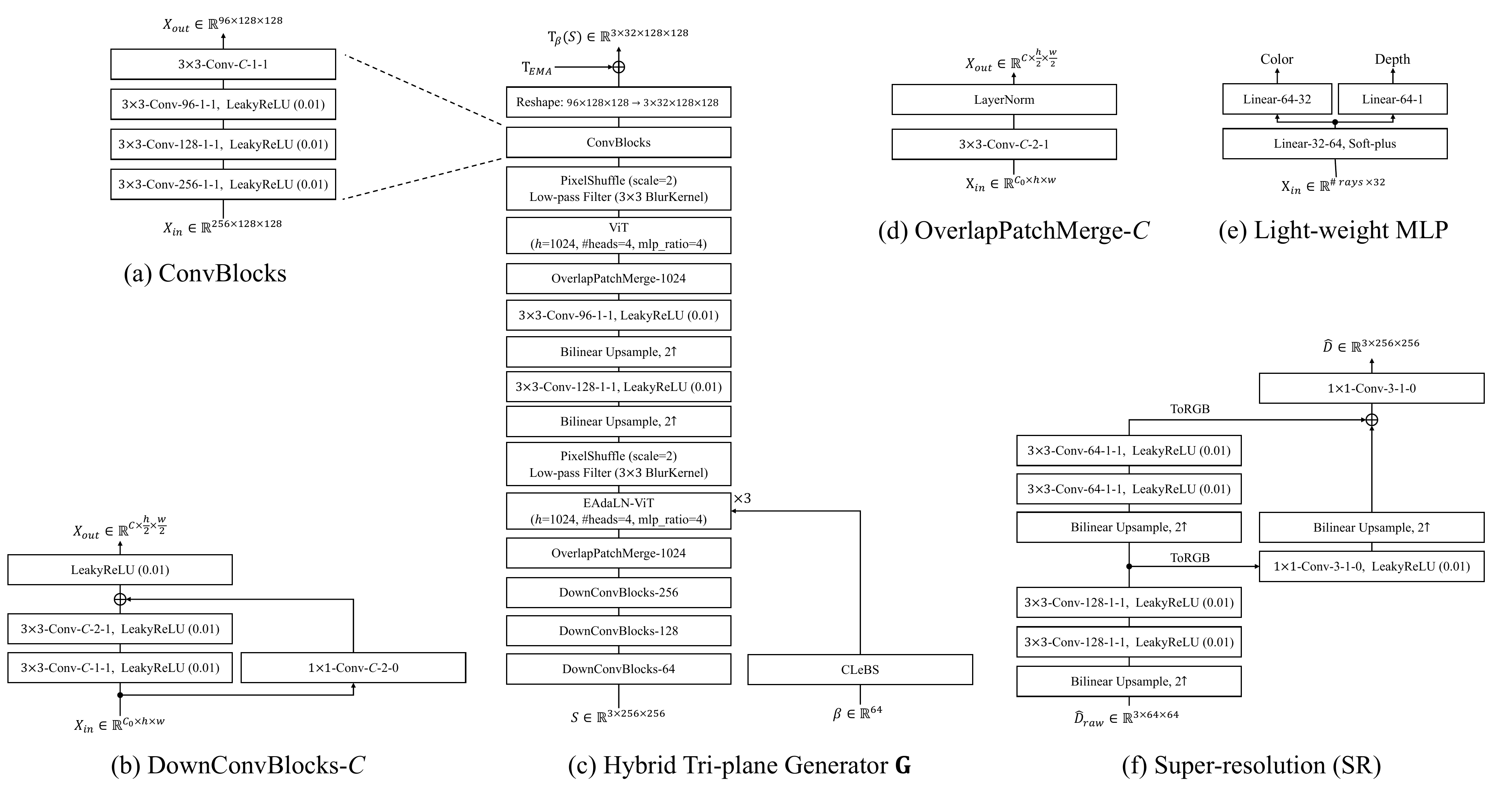}
    \caption{\textbf{The detailed model architectures.} $k \times k$-Conv-$C$-$s$-$p$ is the convolution operator with the kernel size $k\times k$, output channel size $C$, stride step $s$, and padding size $p$. Linear-$C_0$-$C_1$ is the fully-connected layer of the input channel size $C_0$ and the output channel size $C_1$.}  \label{fig:architecture_detail}
    \end{center}
\end{figure}

Our model consists of four parts: Learned Basis Scaling (\textbf{LeBS}), Hybrid Tri-plane Generator $\generator$, Light-weight MLP decoder (\textbf{MLP}) for color and density prediction used in the differentiable volume rendering \cite{nerf}, and Super-resolution (\textbf{SR}) module. The detailed model architectures are shown in \cref{fig:3dmm_vs_lebs} and \cref{fig:architecture_detail}.

\textbf{LeBS} consists of two fully-connected layers along with the learned orthonormal basis $V \subseteq \real^{d}$. We apply QR-decomposition \cite{lia} to a learnable weight in $\real^{d \times n}$ to explicitly compute $V \subseteq \real^{n \times d}$. We set the dimension of the expression space $d = \frac{h}{4}$ to be same as the dimension of the visual tokens where $h=1024$ is the size of the hidden state in the EAdaLN-ViT blocks. We experimentally choose $n = 10$ for the number of basis vectors. We observe that increasing $n$ produces duplicated expression directions. For the contrastive pre-training of LeBS, we employ ResNetSE18 feature extractor \cite{resnetse} followed by a single fully-connected layer to output the $d$-dimensional vector, serving as the image encoder $f_{I}(\cdot)$. Notably, we do not introduce an orthonormal basis to $f_{I}(\cdot)$.

Inspired by \cite{live3d}, we incorporate ViT blocks \cite{vit} into our generator $\generator$, specifically utilizing those from SegFormer \cite{segformer} and DiT \cite{dit}. In both EAdaLN-ViT and ViT, we employ four heads with 1024 hidden dimensions for the multi-head self-attention. It is worth mentioning that the architectures of EAdaLN-ViT and ViT illustrated in \cref{fig:architecture_detail} are the same, with the exception of EAdaLN integration for expression transfer. We employ the exponential moving average (EMA) on the tri-planes for stabilizing the training. More precisely, in the $j$-th gradient step, we calculate and update the EMA $\triplane^j_{EMA}$ and the current tri-plane $\triplane^{j}$ as follows:
\begin{equation}
    \triplane^{j}_{EMA} \leftarrow \delta \cdot \triplane^{j-1}_{EMA} + (1-\delta) \cdot \bar{\triplane}^{j} \quad \text{and} \quad \triplane^{j} \leftarrow \triplane^{j} + \triplane^{j-1}_{EMA} \label{eq:ema}
\end{equation}
where $\bar{\triplane}^{j}$ is the average tri-plane calculated within the $j$-th batch and $\triplane^{0}_{EMA}$ is initialized by $\mathbf{0} \in \real^{3\times32\times128\times128}$. We set $\delta=0.998$ as the weight for the moving average.

\textbf{MLP} for color and density prediction consists of a stack of fully-connected layers with soft-plus activation. In contrast to \cite{eg3d}, we use two fully-connected layers to separately predict them.

For \textbf{SR}, we follow the super-resolution module used in \cite{stylegan2, eg3d} except for the style modulated convolutions.

\subsection{Training Objectives} \label{sec:training}
Our model is trained with reconstruction manner that reconstruct a driving frame $D$ from a source frame $S$ with the driving expression parameters $\beta_{D}$ and camera parameters $p_{D}$ where these frames are randomly sampled from the same video clip. The training consists of two stages. In the first phase, we employ MSE loss $\loss_{2}$ and VGG16 \cite{vgg} multi-scale perceptual loss $\loss_{lpips}$ \cite{lpips} to minimize the perceptual distance between the generated frame $\hat{D}$ and the driving frame $D$. We also minimize the distance between the raw rendered image $\hat{D}_{raw}$ and raw driving image $D_{raw}$ using the same loss functions, denoted by $\loss^{raw}_2$ and $\loss^{raw}_{lpips}$, respectively:
\begin{equation}
    \loss_{rec} = \loss^{raw}_{2} + \loss_{2} + \loss^{raw}_{lpips} + \loss_{lpips}.
    \label{eq:loss_rec}
\end{equation}
In the second phase, we integrate the conditional discriminator used in \cite{stylegan-ada}, using  the camera parameter as additional condition and employing binary cross-entropy loss to compute adversarial loss $\loss_{adv}$. The total loss function $\loss_{total}$ is
\begin{equation}
    \loss_{total} = \lambda_{rec} \loss_{rec} + \lambda_{adv} \loss_{adv},
    \label{eq:loss_total}
\end{equation}
where $\lambda_{rec}$ and $\lambda_{adv}$ are balancing coefficients. 

\subsection{More Implementation Details.}
\subsubsection{Training.} \quad Since our model does not rely on pre-trained EG3D \cite{eg3d, live3d}, it is trained end-to-end, except for CLeBS. For the contrastive pre-training of LeBS, we draw 32 negative samples for each positive sample, set the temperature $\tau$ to 0.07, and train it for 60,000 steps. Longer pre-training does not lead to significant performance improvements.

We empirically set the balancing coefficients in \cref{eq:loss_total} by $\lambda_{rec} = 1$, and $\lambda_{adv}= 0.01$. We train our model for 300,000 steps with the reconstruction loss \cref{eq:loss_rec} and then incorporate the adversarial loss \cref{eq:loss_total} for 10,000 steps to slightly improve the visual quality. For all training, we use Adam \cite{adam} optimizer with the learning rate $10^{-4}$ for $\model$, $10^{-4}$ for CLeBS, and $10^{-5}$ for the discriminator, respectively. Overall training conducts on a single A100 GPU about 5 days with batch size 8. In the inference phase, we use randomly sampled frontal frame as the source frame.

\subsection{Evaluation.}\label{sec:suppl_eval}
\subsubsection{Evaluation metrics.}
\quad We provide additional explanations of the evaluation metrics. Average key-point distance (\textbf{AKD}) is the L1 distance of 68 facial key-points between the generated image and the driving image, which measures the facial structure similarity based on the key-points. We use the face-alignment \cite{face_alignment} to extract the key-points. Cosine similarity of identity embedding (\textbf{CSIM}) is the cosine similarity between the identity embeddings of the source image and the generated image where the embeddings are extracted from ArcFace \cite{arcface}. Average expression distance (\textbf{AED}) and average pose distance (\textbf{APD}) are the L1 distance between the expression parameters (64 dimensions) and the pose parameters (6 dimensions), respectively extracted from the generated image and the driving image. We use the 3DMM extractor \cite{bfm} to extract those parameters.

\begin{figure}[!tb]
\begin{center}
    \captionof{table}{\textbf{Quantitative comparison of on VFHQ with "background".}}\label{tab:vfhq_with_bg}
    \resizebox{0.95\linewidth}{!}{
    \begin{tabular}{l|cccccc|ccc}
    \toprule
    \multicolumn{1}{c|}{\multirow{2}{*}{Method}} & \multicolumn{6}{c|}{Same-identity} & \multicolumn{3}{c}{Cross-identity} \\
        \cline{2-10}
    \multicolumn{1}{c}{} & \multicolumn{1}{|c}{\textbf{PSNR}$\uparrow$} & \multicolumn{1}{c}{\textbf{SSIM}$\uparrow$} & \multicolumn{1}{c}{\textbf{AKD}$\downarrow$} & \multicolumn{1}{c}{\textbf{CSIM}$\uparrow$} & \multicolumn{1}{c}{\textbf{AED}$\downarrow$} & \multicolumn{1}{c}{\textbf{APD}$\downarrow$} & \multicolumn{1}{|c}{\textbf{CSIM}$\uparrow$} & \multicolumn{1}{c}{\textbf{AED}$\downarrow$} & \multicolumn{1}{c}{\textbf{APD}$\downarrow$} \\
        \hline
        ROME & 8.309 & 0.400 & 11.179 & 0.592 & 0.123 & 0.173  & 0.495 & 0.236 & 0.201  \\
        OTAvatar & 10.667 & 0.457 & 15.236 & 0.492 & 0.181 & 0.182  & 0.492 & 0.288 & 0.237\\        
        HiDeNeRF & 12.254 & 0.345 & 22.136 & 0.354 & 0.135 & 0.252  & 0.408 & 0.259 & 0.230 \\ 
        \hline
        \textbf{Ours} & \textbf{23.555} & \textbf{0.704} & \textbf{3.453} & \textbf{0.811} & \textbf{0.082} & \textbf{0.030}  & \textbf{0.694} & \textbf{0.208} & \textbf{0.080}\\
        \bottomrule
    \end{tabular}}
    \end{center}
\end{figure}

\begin{figure}[!tb]
    \begin{center}
        \begin{subfigure}[t]{0.485\textwidth}
            \includegraphics[width=1\textwidth]{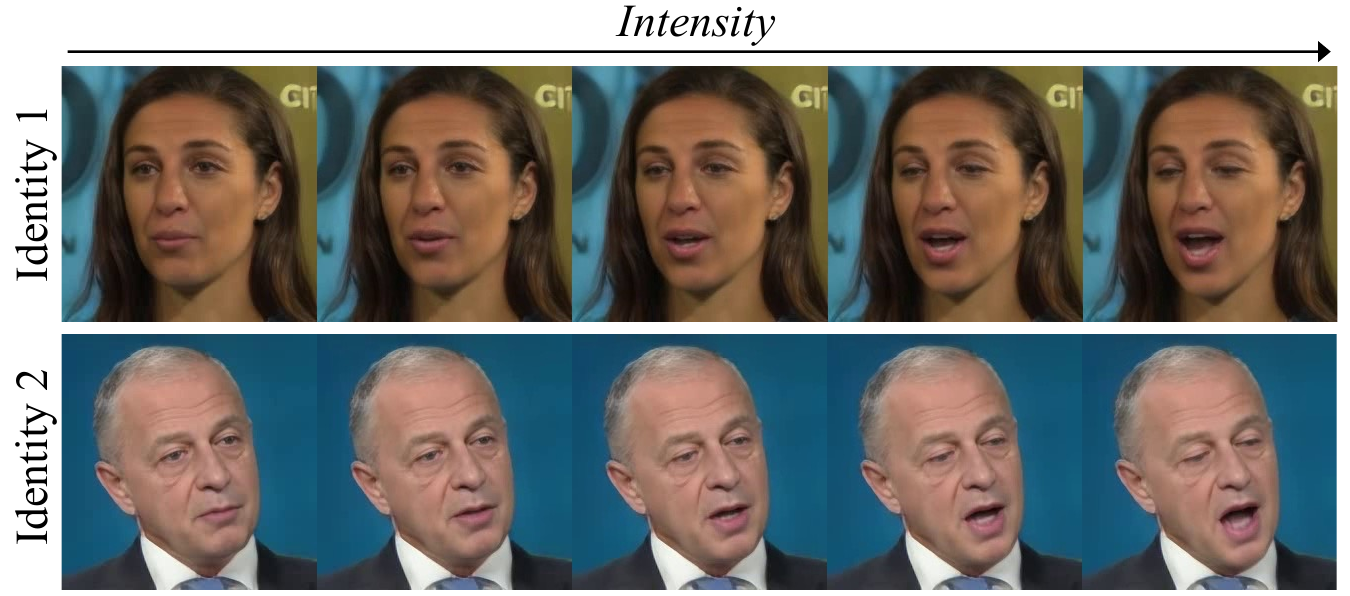} 
            \caption{Basis vector $\mathbf{v}_{1}$}
            \label{fig:linear_c}
        \end{subfigure}
        \begin{subfigure}[t]{0.485\textwidth}
            \includegraphics[width=1\textwidth]{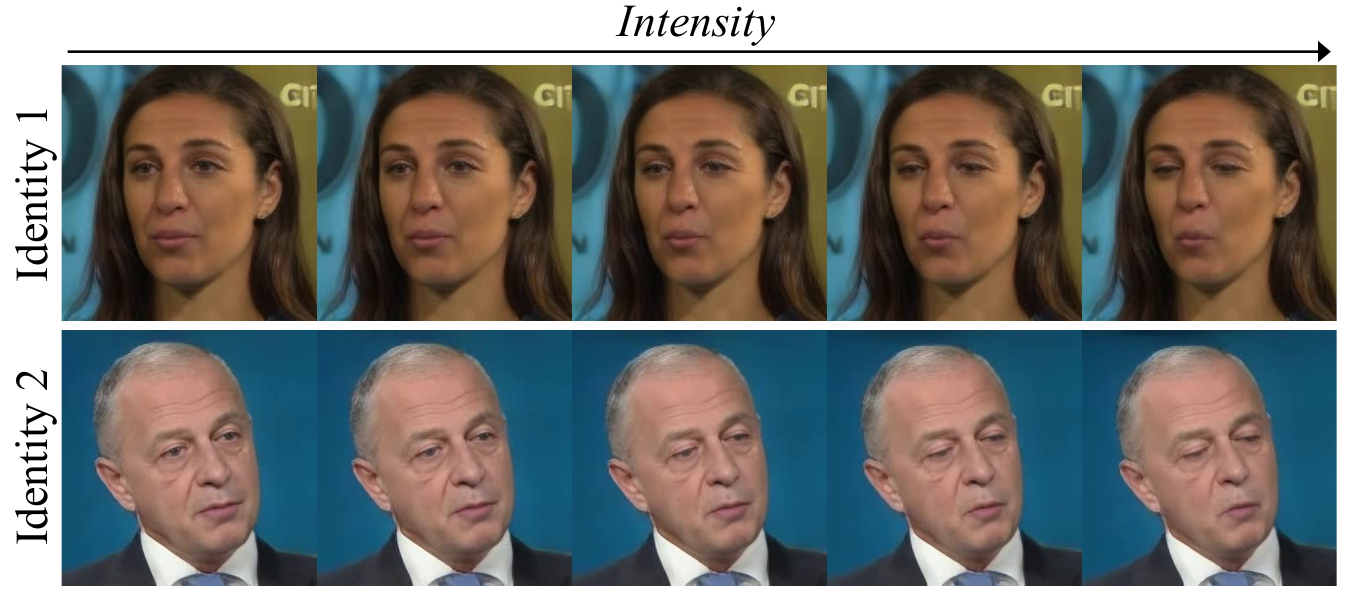}
            \caption{Basis vector $\mathbf{v}_3$}
            \label{fig:linear_d}
        \end{subfigure}
        \begin{subfigure}[t]{0.485\textwidth}
            \includegraphics[width=1\textwidth]{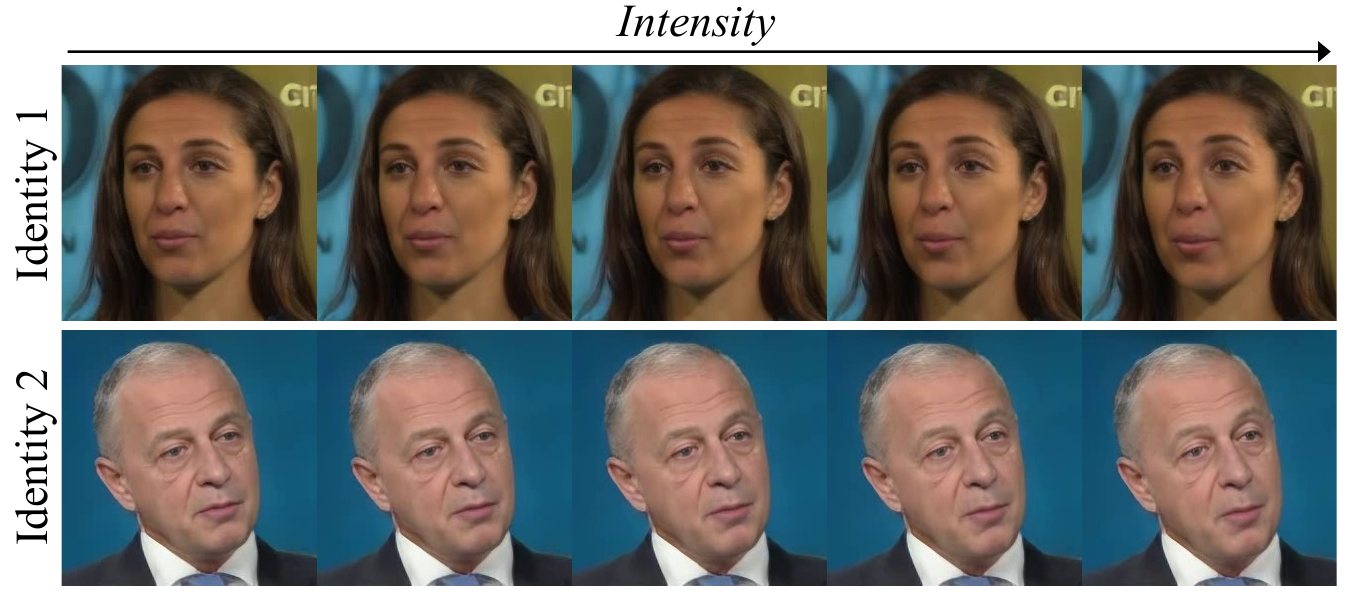}
            \caption{Basis vector $\mathbf{v}_6$}
            \label{fig:linear_f}
        \end{subfigure}
        \begin{subfigure}[t]{0.485\textwidth}
            \includegraphics[width=1\textwidth]{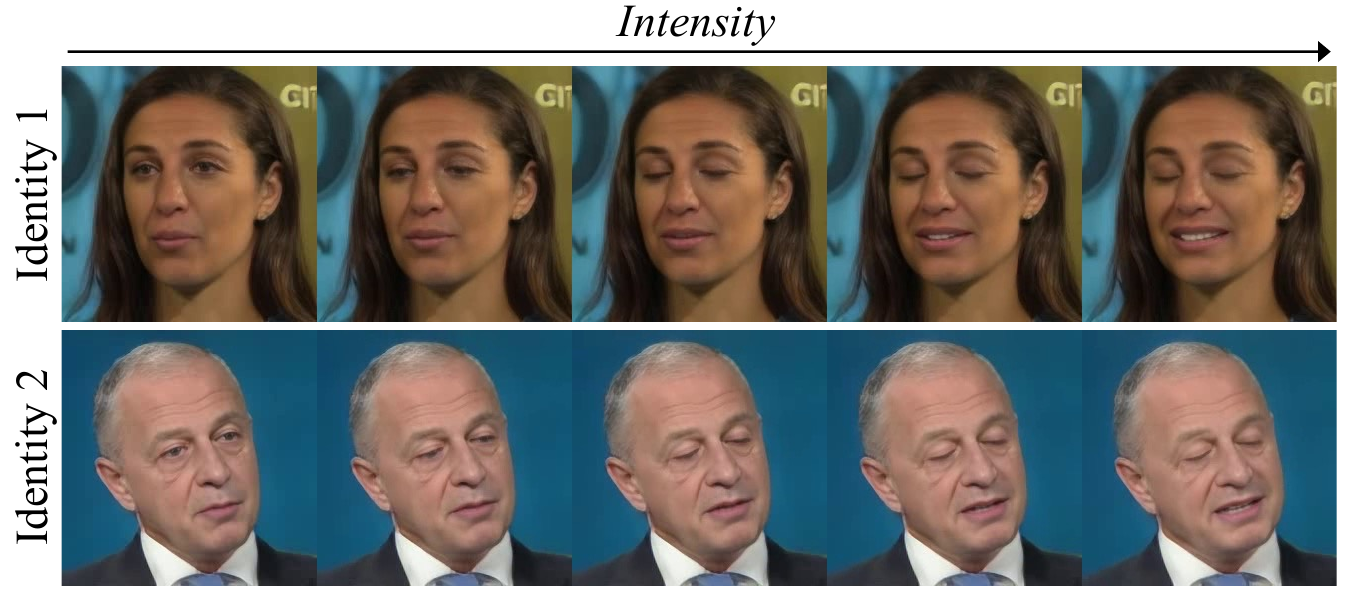} 
            \caption{Basis vector $\mathbf{v}_{9}$}
            \label{fig:linear_e}
        \end{subfigure}
    \caption{\textbf{Linear scaling along the different basis vectors of CLeBS.}} \label{fig:linear_supple}
    \end{center}
\end{figure}

\subsection{Additional Results.} \label{sec:suppl_additional_qualitative}
\subsubsection{Further comparison without removing the background.} \quad In \cref{tab:vfhq_with_bg}, we provide additional quantitative comparison with ROME \cite{rome}, OTAvatar \cite{otavatar}, and HiDe-NeRF \cite{hidenerf} to verify that these models have advantage on the evaluation metrics without background.

\subsubsection{Linear scaling along the orthonormal basis.}
\quad In \cref{fig:linear_supple}, we show additional results of linear scaling along the different basis vectors \cite{lia}. For $\mathbf{v}_1$, we scale $\lambda_1$ from 1 to -7, showing mouth opening and eye closing. For $\mathbf{v}_3$, we scale $\lambda_3$ from 1 to 20, showing eye closing and lip pursing. For $\mathbf{v}_6$, we scale $\lambda_{6}$ from 1 to -7, showing eyebrow moving. For $\mathbf{v}_{9}$, we scale $\lambda_{9}$ 1 from to -10, showing eye closing and smiling. Since our method does not constrain the range of the coefficients $\lambda=(\lambda_i)_{i=1}^{10}$, the manipulation can be realized along the negative scaling. Please refer to video results.

\begin{figure}[!t]
    \begin{center}
        \includegraphics[width=\textwidth]{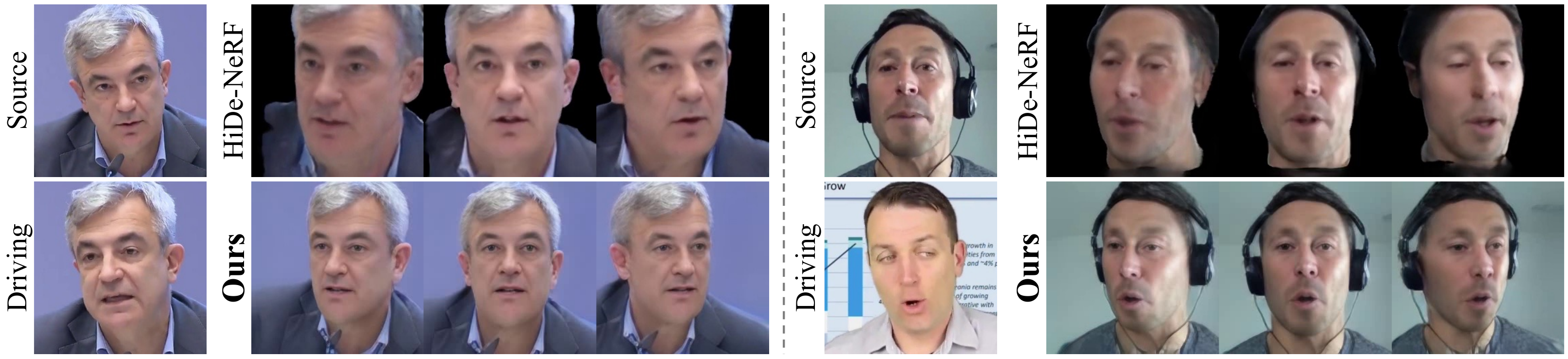}
    \caption{\textbf{Novel-view synthesis results with expression transfer.}} \label{fig:vs_hidenerf_supple}
    \end{center}
\end{figure}
\begin{figure}[!t]
    \begin{center}
        \includegraphics[width=\textwidth]{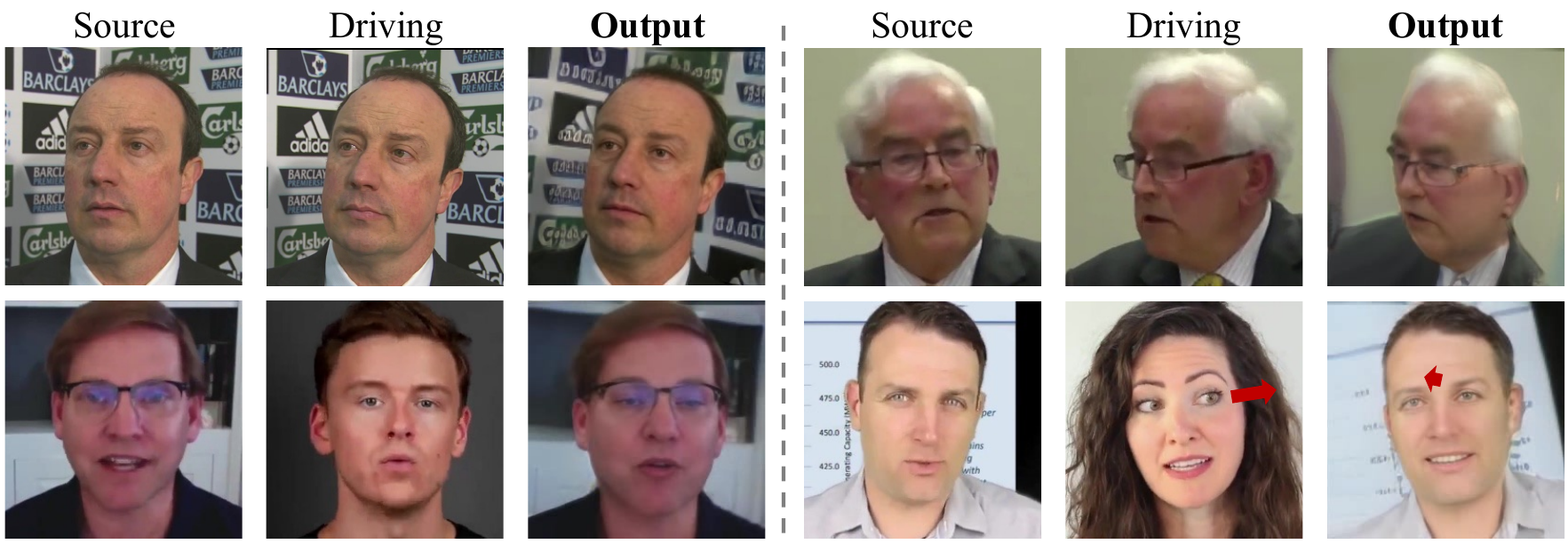}
    \caption{\textbf{Limitation cases of $\model$}. The red arrows indicate the directions of eye gaze.} \label{fig:limitations}
    \end{center}
\end{figure}
\subsubsection{Additional comparison with HiDe-NeRF.} \quad In \cref{fig:vs_hidenerf_supple}, we exhibit additional comparison results with HiDe-NeRF \cite{hidenerf} for novel-view synthesis with expression transfer. Please refer to the video results for further details. 

\subsection{Limitations and Future Work.} \label{sec:failure}
We exhibit the limitation cases of $\model$ in \cref{fig:limitations}. Since the tri-plane represents \cite{eg3d} the foreground and the background as a whole, our model jointly renders them, resulting in head pose-aligned distortion. Several prior works \cite{goavatar, hidenerf, rome, otavatar} address this issue by removing the complex background and providing the volume rendering with a uniform background. However, they heavily rely on the performance of the background segmentation model \cite{modnet}, exhibiting the temporal jitters in the generated videos. Additionally, our model cannot control eye gazing since the 3DMM parameters do not model eye movement.
We leave these limitations for future research.

\end{document}